\documentclass{article} %
\usepackage{conference,times}
\usepackage{subcaption}
\usepackage{multirow}

\usepackage{amsmath,amsfonts,bm}

\def\eqref#1{equation~\ref{#1}}

\def\1{\bm{1}}

\def\rvepsilon{{\mathbf{\epsilon}}}

\def\rvh{{\mathbf{h}}}

\def\rvv{{\mathbf{v}}}

\def\rvx{{\mathbf{x}}}
\def\rvy{{\mathbf{y}}}

\DeclareMathAlphabet{\mathsfit}{\encodingdefault}{\sfdefault}{m}{sl}
\SetMathAlphabet{\mathsfit}{bold}{\encodingdefault}{\sfdefault}{bx}{n}

\usepackage{hyperref}
\usepackage{booktabs}
\usepackage{array} 
\usepackage{url}
\usepackage{graphicx}
\usepackage{wrapfig}
\usepackage{needspace}
\usepackage{float}
\usepackage{booktabs}
\usepackage[table]{xcolor}
\usepackage{enumitem}
\usepackage{caption}
\usepackage{arydshln}
\usepackage{siunitx}
\usepackage{makecell}
\usepackage[table]{xcolor}
\definecolor{cornellred}{rgb}{0.7, 0.11, 0.11}
\definecolor{cadmiumgreen}{rgb}{0.0, 0.42, 0.24}
\definecolor{aliceblue}{rgb}{0.91, 0.94, 0.97}
\definecolor{darkblue}{rgb}{0.83, 0.89, 0.97}
\definecolor{Red7}{rgb}{0.941, 0.243, 0.243}
\definecolor{Green7}{RGB}{55, 178, 77}
\definecolor{Blue9}{rgb}{0.098,0.3,0.9}
\hypersetup{  
linkcolor = cornellred,  
citecolor  = cadmiumgreen,  
colorlinks = true,  
urlcolor = Blue9
}
\title{Scaffold Then Internalize: Representation Injection for Diffusion Transformers}
\author{Han Fu$^{1,2}$\quad Jiacheng Chen$^1$\quad Baoquan Zhao$^1$\\
\textbf{Weidong Chen$^2$\quad Wei Liu$^2$\quad Qing Li$^3$\quad Xudong Mao$^1$\thanks{Corresponding author.}}\\[3pt]
\normalfont $^1$Sun Yat-sen University\quad $^2$Video Rebirth\quad $^3$The Hong Kong Polytechnic University
}

\iclrfinalcopy %
\begin{document}

\maketitle
\begin{abstract}
Recent representation alignment (REPA) methods accelerate diffusion transformer training by aligning projections of the transformer's hidden states with representations from pretrained visual encoders. In this work, we explore a reverse and complementary direction to REPA: rather than projecting diffusion representations into the encoder's space, we inject encoder representations into the diffusion transformer, allowing them to actively participate in the denoising process. To this end, we introduce \textit{REPresentation Injection} (REPI), a training framework based on a scaffold-to-internalization strategy, in which projected encoder representations initially serve as a temporary scaffold and are then progressively internalized by the diffusion transformer. REPI outperforms REPA across a wide range of backbones and is highly complementary to it: combining the two yields substantial gains over either alone. Notably, with only 160K training steps, REPI + REPA matches vanilla SiT trained for 7M steps, a speedup of over $43.5\times$. Code will be available at \url{https://jeneveuxpas.github.io/REPI}.
\end{abstract}

\begin{figure*}[ht!]
 \vspace{-0.0in}
    \centering
    \includegraphics[width=\linewidth]{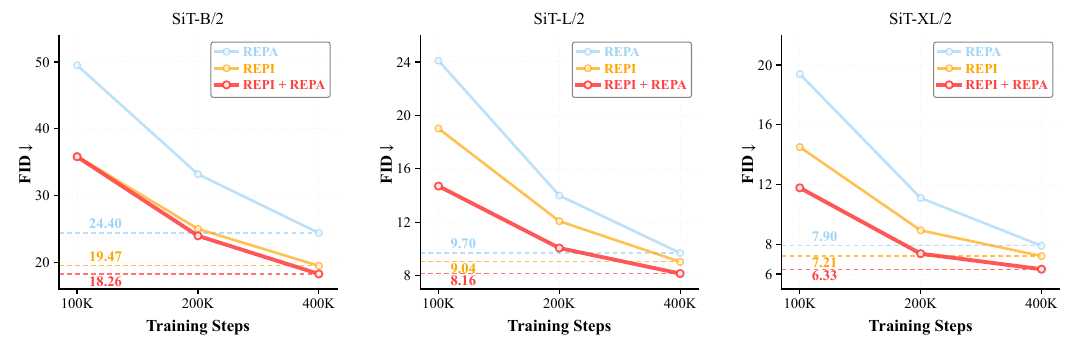}
    \caption{
    \textbf{REPI accelerates diffusion transformer training across model scales.} Compared with REPA, REPI consistently achieves lower FID on SiT-B, SiT-L, and SiT-XL, and combining the two yields further gains. Notably, on SiT-XL, REPI + REPA at 200K training steps already outperforms REPA trained for 400K steps.
    }
    \label{fig:teaser}
\end{figure*}

\section{Introduction}

Diffusion Transformers (DiTs) \citep{dit,sit,sd3} exhibit strong scalability for visual generation, but achieving high sample quality still requires extensive training. Representation Alignment (REPA) \citep{repa} attributes this inefficiency to the difficulty DiTs face in learning semantically structured visual representations, and addresses it by distilling knowledge from a pretrained visual encoder: the diffusion transformer's hidden states are projected and aligned with the encoder's representations via an auxiliary objective \citep{irepa,wang2025repa}. In this paper, we explore a reverse and complementary direction: \textit{rather than projecting diffusion representations into the encoder's space, can encoder representations instead be projected into the diffusion transformer and actively participate in denoising during training?}

We begin with a naive design that directly reverses REPA: the encoder output is projected through a lightweight projection layer, and the resulting projection entirely replaces the hidden state at an intermediate layer of the diffusion transformer. However, since this hidden state is the output of a transformer block (see Figure~\ref{fig:injection_interfaces} for an illustration), overwriting it discards all prior computation on the noisy input. As a result, subsequent blocks can no longer condition their predictions on the noisy input, causing training to collapse and generated samples to remain pure noise.

This observation motivates moving the injection point inside the transformer block: rather than overwriting the block output, we replace an intermediate quantity within the block, such as the attention output or the key/value (KV) representations. This preserves the flow of information from the noisy input to the block output. For instance, replacing the KV representations still allows the preceding computation to reach subsequent blocks through the query (Q) representations and the residual shortcut. To investigate the potential of this injection strategy, we first consider an idealized oracle setting with access to encoder representations at inference time. Within only 30K training steps, this oracle injection method achieves a lower FID than vanilla SiT \citep{sit} trained for 7M steps, indicating that encoder representations can provide a useful signal for denoising.

However, this oracle setting relies on encoder representations extracted from clean images at inference time, which are unavailable in standard generation. To close this gap, we introduce \textit{REPresentation Injection} (REPI), which adopts a scaffold-to-internalization training strategy that eliminates the need for encoder representations at inference time. REPI treats the injected encoder representations as a scaffold: during early training, the internal representations at one layer of the diffusion transformer are replaced by projections of the encoder representations; the scaffold is then removed, and the diffusion transformer resumes computing its own representations at that layer. To facilitate this internalization, we introduce an internalization objective that encourages the self-computed representations to match the injected encoder representations. This process allows the model to first learn to exploit the semantically structured representations provided by the scaffold, and then to learn to reproduce compatible representations on its own. After training, the visual encoder is discarded, and the resulting model is architecturally identical to the original diffusion transformer at inference.

\textbf{Complementarity with REPA.} REPI is highly complementary to REPA, as the two methods leverage encoder representations through different mechanisms: REPA encourages the diffusion transformer to learn the structure of encoder representations through an auxiliary alignment loss, while REPI injects and internalizes these representations into the denoising computation, adapting their structure to the denoising task. Empirically, on SiT-XL \citep{sit}, REPI alone achieves an FID of 14.51 at 100K training steps, compared with 19.40 for REPA, and combining both methods further reduces FID to 11.78, substantially outperforming either method alone.

We conduct extensive experiments on class-conditional ImageNet \citep{imgnet} generation across multiple model scales to demonstrate the effectiveness of our method. Our approach achieves superior generation quality and training efficiency compared with several state-of-the-art baselines. Notably, on SiT-XL, REPI combined with REPA achieves an FID of 8.22 within only 160K training steps, matching the FID of 8.30 obtained by vanilla SiT after 7M steps, a training speedup of over $43.5\times$. Furthermore, our method at 200K steps already outperforms REPA trained for 400K steps.

Our main contributions are summarized as follows:
\vspace{-0.05in}
\begin{itemize}[leftmargin=5mm,itemsep=7pt,parsep=0pt,topsep=7pt,partopsep=0pt]
    \item We explore a direction that is reverse and complementary to REPA: rather than projecting diffusion representations into the encoder's space, we inject encoder representations into the diffusion transformer, allowing them to actively participate in the denoising process.
    \item We introduce REPI, a training framework that internalizes semantically structured representations from pretrained visual encoders into the diffusion transformer via a scaffold-to-internalization strategy.
    \item We show that REPI alone outperforms REPA, while the two remain highly complementary. Combined with REPA, our model trained for only 160K steps surpasses vanilla SiT trained for 7M steps, a speedup of over $43.5\times$.
\end{itemize}

\section{Preliminaries}
\label{sec:preliminaries}

\textbf{Diffusion with flow matching.}
Given a clean sample $\rvx_* \sim p_{\text{data}}$ and condition $c$, flow matching~\citep{fmgen,albergo2022building} constructs
$\rvx_t=\alpha_t \rvx_* + \sigma_t \rvepsilon$ on $t\in[0,1]$, where $\rvepsilon\sim\mathcal{N}(0,I)$, $(\alpha_0,\sigma_0)=(1,0)$, and $(\alpha_1,\sigma_1)=(0,1)$. A denoiser $\rvv_\theta(\rvx_t,t,c)$ is trained by
\begin{equation}
\mathcal{L}_{\text{velocity}}(\theta)
    = \mathbb{E}_{\rvx_*,\,\rvepsilon,\,t}\Bigl[\,\bigl\lVert
      \rvv_\theta(\rvx_t,t,c) - \rvv_t \bigr\rVert^2\,\Bigr],
\qquad
\rvv_t=\dot{\alpha}_t\rvx_*+\dot{\sigma}_t\rvepsilon.
\label{eq:diffusion}
\end{equation}

\vspace{-5pt}
\textbf{Representation alignment.}
REPA~\citep{repa} aligns an intermediate denoiser state $\rvh_t$ with clean-image features $\rvy_*=f(\rvx_*)$ from a pretrained visual encoder $f$. A projection head $h_\phi$ maps $\rvh_t$ into the encoder space, yielding
\begin{equation}
\mathcal{L}_{\text{REPA}}(\theta,\phi) = -\mathbb{E}_{\rvx_*,\rvepsilon,t}\Bigl[\frac{1}{N}\sum_{n=1}^{N}\text{sim}\bigl(h_\phi(\rvh_t^{[n]}),\, \rvy_*^{[n]}\bigr)\Bigr],
\end{equation}
where $n$ indexes patches and $\mathrm{sim}(\cdot,\cdot)$ is a predefined similarity function.

\section{Can Encoder Representations Participate in Denoising?}
\label{sec:oracle}
REPA transfers knowledge from a pretrained visual encoder to a diffusion transformer through a regularization term that aligns the transformer's projected hidden states with target representations from the encoder. In this section, we investigate a complementary question: rather than aligning with encoder representations, can they instead be mapped directly into the diffusion transformer and participate actively in denoising? We first examine this question under a diagnostic oracle setting, in which encoder representations are assumed to be available at inference time. Section~\ref{sec:method} then introduces our method, which eliminates the need for encoder representations at inference time.

\begin{wraptable}{r}{0.25\textwidth}
\vspace{-13pt}
\centering
\small
\setlength{\tabcolsep}{5pt}
\caption{Oracle results (SiT-XL/2, 30K steps).}
\vspace{-8pt}
\label{tab:oracle}

\resizebox{1\linewidth}{!}{%
\begin{tabular}{lr}
\toprule
Injection target & FID$\downarrow$ \\
\midrule
Hidden state & 212.5 \\
Attention output & 5.0 \\
Q/K/V & 5.2 \\
K/V & 7.1 \\
K only & 12.4 \\
V only & 4.6 \\
\bottomrule
\end{tabular}%
}

\vspace{-8pt}
\end{wraptable}

A naive design is to simply reverse the mapping direction of REPA: instead of projecting diffusion hidden states into the encoder's space, we use a lightweight projection layer to map encoder representations into the diffusion hidden-state space and substitute them for an intermediate hidden state. This design fails catastrophically, yielding an FID of 212.5 (Table~\ref{tab:oracle}). We attribute this failure to the fact that the hidden state is the output of a transformer block (Figure~\ref{fig:injection_interfaces}), so replacing it entirely severs all subsequent blocks from the noisy input. This failure suggests that external representations must be injected without overwriting the input-dependent stream. We therefore instead inject representations inside the attention module, testing five alternatives that all preserve the flow of information from the noisy input to the block output (Table~\ref{tab:oracle}). Details of these variants are provided in Appendix~\ref{app:oracle}.

As shown in Table~\ref{tab:oracle}, all five variants achieve strong oracle FID after only 30K training steps, with V-only injection performing best. A plausible explanation is that V-only injection preserves the diffusion transformer's native attention routing through Q and K, while injecting semantic content from the encoder via V. Note that the optimal injection location differs for our full model, as discussed in Section~\ref{sec:injection}.

Although this oracle setting is impractical, as it assumes access to encoder representations at inference time, these results indicate that projected encoder representations can substantially benefit the denoising process. We next describe how our method eliminates this dependence on encoder representations.

\begin{figure}[t]
  \centering
  \begin{subfigure}[t]{0.465\textwidth}
    \centering
    \includegraphics[width=\linewidth]{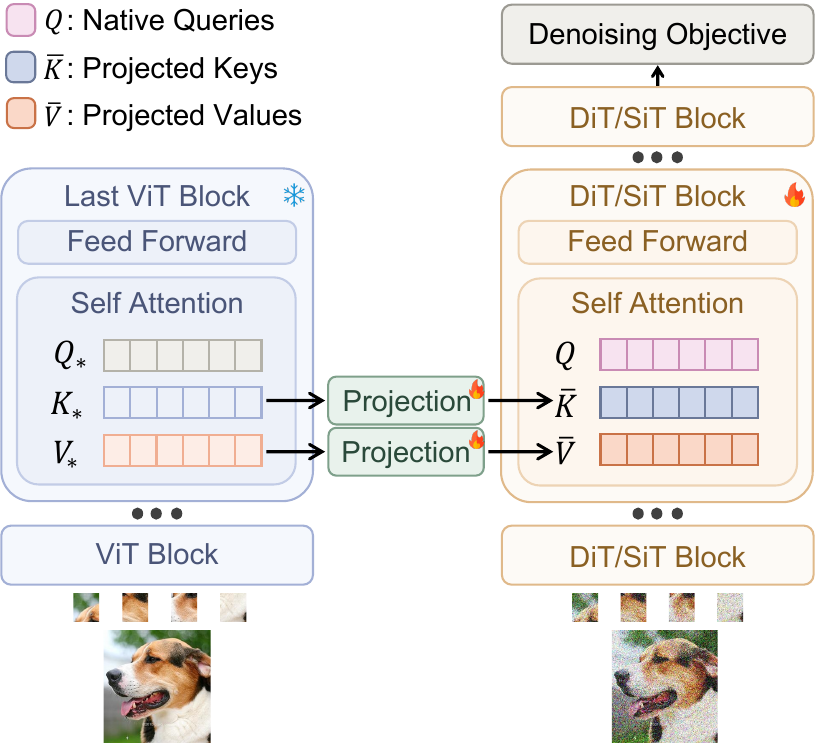}
    \caption{Scaffold}
    \label{fig:framework_a}
  \end{subfigure}
  \hspace{0.05\textwidth}
  \begin{subfigure}[t]{0.465\textwidth}
    \centering
    \includegraphics[width=\linewidth]{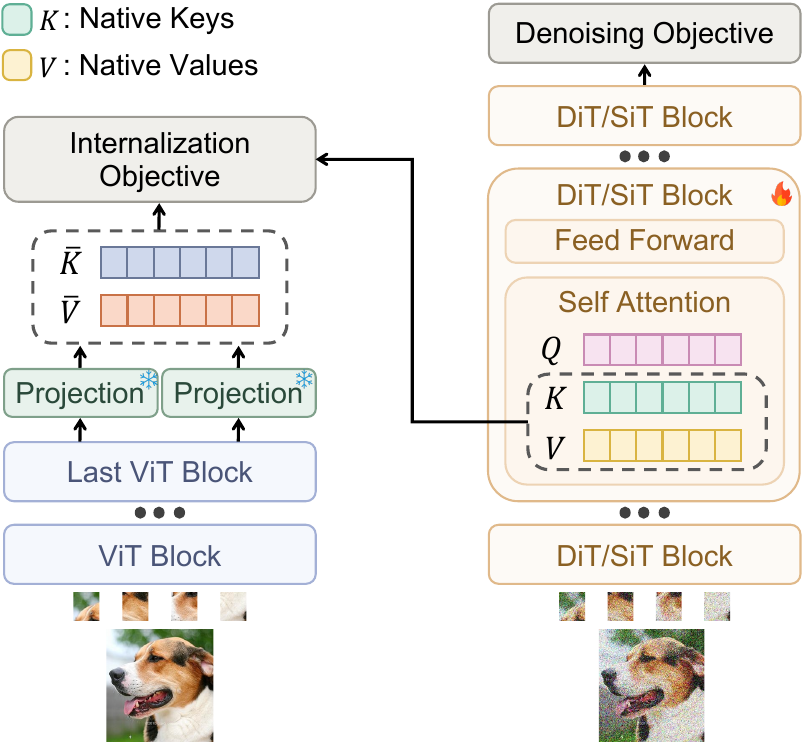}
    \caption{Internalization}
    \label{fig:framework_b}
  \end{subfigure}
  \caption{\textbf{Overview of REPI.}
  (a) \textit{Scaffold}: projected keys and values from a frozen visual encoder temporarily replace the diffusion transformer's native keys and values, while its native queries are preserved.
(b) \textit{Internalization}: the model resumes using its own keys and values, guided by an internalization objective that aligns them with the projected encoder representations. At inference time, the encoder and projection layers are discarded.
  }
  \label{fig:framework}
  \vspace{-3pt}
\end{figure}

\section{REPI: Representation Injection and Internalization}
\label{sec:method}

\subsection{Overview}
\label{sec:overview}

The oracle setting in Section~\ref{sec:oracle} requires representations extracted from clean images during sampling, which are unavailable in standard generation. To eliminate this dependency, we propose \textit{REPresentation Injection} (REPI), which treats projected encoder representations as a scaffold and internalizes their benefit into the diffusion transformer. As illustrated in Figure~\ref{fig:framework}, REPI projects a representation extracted from a clean image and substitutes it for the native representation within the diffusion transformer (Section~\ref{sec:injection}). After a small number of training steps, we remove the scaffold and restore the native computation, while an internalization objective encourages the native representation to reproduce the representation previously supplied by the scaffold (Section~\ref{sec:internalization}). At inference time, the visual encoder and projection layers are discarded, and sampling relies solely on the original diffusion backbone.

\subsection{Representation Injection}
\label{sec:injection}

REPI initially substitutes projections of clean-image encoder representations for the native representations at a selected diffusion layer, serving as a temporary scaffold that is later removed as training progresses. Specifically, we inject K/V representations, as we empirically find this choice achieves the best results (Table~\ref{tab:ablation_injection_target}). This finding is inconsistent with the oracle results in Section~\ref{sec:oracle}, where V-only injection performs best. One possible explanation is that the oracle setting only needs to consider how effectively an external representation can be consumed, whereas REPI additionally requires that the injected representation be subsequently internalized and removed.

Formally, let $f$ denote a pretrained visual encoder and $\rvx_*$ a clean image. During the forward pass of $f(\rvx_*)$, we extract the key and value representations $(K_*, V_*)$ computed within its final self-attention layer. At a selected self-attention layer of the diffusion transformer, we introduce two trainable projection layers, $h_{\phi_K}$ and $h_{\phi_V}$, which map $K_*$ and $V_*$ into the corresponding key and value spaces of that layer:
\begin{equation}
    \bar{K} = h_{\phi_K}(K_*), \qquad
    \bar{V} = h_{\phi_V}(V_*).
\end{equation}
We then replace the native key and value representations with $\bar{K}$ and $\bar{V}$, respectively. The resulting attention output is
\begin{equation}
    Z_{\mathrm{inj}}
    =
    \operatorname{softmax}
    \left(
        \frac{Q\bar{K}^\top}{\sqrt{d}}
    \right)\bar{V},
    \label{eq:repi_injection}
\end{equation}
where $Q$ denotes the native query representation and $d$ is the query/key dimension. For clarity, we omit layer and attention-head indices. REPI changes only the source of the keys and values, leaving the other operations of the attention block unchanged. Notably, in addition to softmax attention, this substitution generalizes to the gated linear attention used in DiG~\citep{DiG}, as demonstrated in Figure~\ref{fig:generality}.

This design allows information from the noisy input to continue propagating through the native query $Q$ and the residual shortcut, preserving the input-dependent information accumulated by preceding blocks. Meanwhile, the projections $\bar{K}$ and $\bar{V}$ provide a semantically structured key-value memory derived from the clean-image representation, in which the keys determine which external information is retrieved, while the values determine its content.

In practice, we implement $h_{\phi_K}$ and $h_{\phi_V}$ as linear layers, and select the same diffusion transformer layer and encoder layer as REPA. During training, the diffusion transformer and the projection layers are jointly optimized using the diffusion objective (Eq.~\ref{eq:diffusion}).

\subsection{Internalization and Scaffold Removal}
\label{sec:internalization}
The injected encoder representations provide semantically organized content for denoising, but cannot be retained during standard sampling, as their construction requires a clean image. We therefore treat these injected representations not as permanent conditioning, but as a temporary representation scaffold, which we later remove while internalizing its benefit into the diffusion transformer's native representations.

A simple strategy is to abruptly remove the scaffold and restore the diffusion transformer's native representations after a small number of training steps. Surprisingly, even this naive switch improves FID from 39.4 for vanilla SiT to 22.45 at 100K steps. We hypothesize that the scaffold provides a favorable initialization for the other transformer layers: while the scaffold is active, these layers learn to exploit semantically structured encoder representations for denoising. After the scaffold is removed, this learned capability may facilitate subsequent training as the model adapts to its own native representations.

To internalize the encoder representations more thoroughly, we introduce an internalization loss that explicitly encourages the native K/V representations to reproduce their encoder-derived counterparts. Specifically, the native representations $(K, V)$ are trained to match the projections $(\bar{K}, \bar{V})$:
\begin{equation}
    \mathcal{L}_{\mathrm{int}}
    =
    \frac{1}{|K|}
    \left\|K-\operatorname{sg}\!\left[\bar{K}\right]\right\|_F^2
    +
    \frac{1}{|V|}
    \left\|V-\operatorname{sg}\!\left[\bar{V}\right]\right\|_F^2,
    \label{eq:internalization}
\end{equation}
where $\operatorname{sg}[\cdot]$ denotes the stop-gradient operator and $|\cdot|$ denotes the number of elements in the corresponding representation.

This internalization loss also facilitates a smooth transition, as it encourages the native K/V representations to remain consistent with their pre-transition states. We additionally experimented with a linearly scheduled transition but observed no further gain, since the internalization loss alone already renders the switch sufficiently smooth.

In practice, this internalization term is added to the diffusion loss, yielding:
\begin{equation}
    \mathcal{L}
    =
    \mathcal{L}_{\mathrm{velocity}}
    +
    \lambda\mathcal{L}_{\mathrm{int}},
    \label{eq:repi_objective}
\end{equation}
where $\lambda$ controls the tradeoff between denoising and internalization.

\subsection{Combining REPI with REPA}
\label{sec:combining}

We empirically find that REPI and REPA are highly complementary (Figure~\ref{fig:teaser}), as the two methods leverage encoder representations through fundamentally different mechanisms. REPA aligns projections of intermediate diffusion representations with corresponding encoder representations, providing diffusion-to-encoder alignment supervision. In contrast, REPI injects and internalizes projections of encoder representations into the denoising computation, forming an encoder-to-diffusion injection mechanism.

When combining the two methods, REPA supervision is applied throughout the entire training process, from scaffold to internalization. We place the REPI injection layer before the REPA alignment layer, an ordering that offers two benefits. First, it allows the two mechanisms to operate cooperatively rather than interfere with each other: injecting after the alignment layer would instead alter the subsequent computation, weakening the influence of the REPA supervision applied earlier. Second, once the scaffold is removed, REPA continues to provide semantic supervision at the alignment layer, complementing the internalization objective. Table~\ref{tab:repi_repa_depth} reports results for different injection-layer choices, showing that reversing the order still improves performance, but the gains are substantially smaller.

\begin{table*}[t]
\centering
\footnotesize
\noindent

\colorlet{repirow}{gray!8}
\colorlet{combinedrow}{gray!14}

\begin{minipage}[t]{0.3415\textwidth}
\vspace{0pt}
\caption{\textbf{FID comparison on SiTs.}
All results are reported on ImageNet $256\times256$ without CFG.}
\label{tab:fid_comparison}
\end{minipage}
\hfill
\begin{minipage}[t]{0.6385\textwidth}
\vspace{0pt}
\caption{\textbf{\boldmath Quantitative comparison on ImageNet $256\times256$ with CFG}.
Asterisks (*) denote CFG scheduling with the guidance interval~\citep{kynkaanniemi2024applying}.}
\label{tab:cfg_comparison}
\end{minipage}

\par\nointerlineskip
\vspace{-6pt}

\noindent
\begin{minipage}[t]{0.3415\textwidth}
\vspace{0pt}
\centering
\begingroup
\renewcommand{\arraystretch}{1.05}

\setlength{\tabcolsep}{5.5pt}
\resizebox{\linewidth}{!}{%
\renewcommand{\arraystretch}{0.956}
\begin{tabular}{
    lc
    S[
        table-format=2.2,
        detect-weight=true,
        mode=text
    ]
}
    \toprule
    Method & Iter. & {FID$\downarrow$} \\
    \midrule

    SiT-B/2                        & 400K & 33.00 \\
    + REPA                         & 400K & 24.40 \\
    \rowcolor{repirow}
    \textbf{+ REPI (ours)}         & \bfseries 400K & \bfseries 19.47 \\
    
    \midrule
    
    SiT-L/2                        & 400K & 18.81 \\
    + REPA                         & 400K & 9.70 \\
    \rowcolor{repirow}
    \textbf{+ REPI (ours)}         & \bfseries 400K & \bfseries 9.04 \\
    
    \midrule
    
    SiT-XL/2                       & 7M   & 8.30  \\
    + REPA                         & 100K & 19.40 \\
    + REPA                         & 400K & 7.90  \\
    + iREPA                        & 100K & 16.96 \\
    + iREPA                        & 400K & 7.52  \\
    + sREPA                        & 100K & 15.40  \\
    + sREPA                        & 400K & 7.17  \\
    + Stable Velocity              & 100K & 17.12 \\
    + Stable Velocity              & 400K & 7.58  \\
    
    \addlinespace[1pt]
    
    \rowcolor{repirow}
    \textbf{+ REPI (ours)}         & 100K & 14.51 \\
    \rowcolor{repirow}
    \textbf{+ REPI (ours)}         & 400K & 7.21 \\
    
    \rowcolor{combinedrow}
    \textbf{+ REPA + REPI (ours)}  &\bfseries 100K &\bfseries 11.78 \\
    \rowcolor{combinedrow}
    \textbf{+ REPA + REPI (ours)}  & 160K & 8.22  \\
    \rowcolor{combinedrow}
    \textbf{+ REPA + REPI (ours)}  &\bfseries 400K & \bfseries 6.33 \\
    \bottomrule
\end{tabular}
}

\endgroup
\end{minipage}
\hfill
\begin{minipage}[t]{0.6385\textwidth}
\vspace{0pt}
\centering
\begingroup
\renewcommand{\arraystretch}{1.015}
\centering
\resizebox{\linewidth}{!}{%
\begin{tabular}{lcccccc}
\toprule
Model & Epochs & FID$\downarrow$ & sFID$\downarrow$ & IS$\uparrow$ & Prec.$\uparrow$ & Rec.$\uparrow$ \\
\midrule
\multicolumn{7}{l}{\textit{Latent Diffusion Transformer}} \\
MaskDiT         & 1600 & 2.28 & 5.67 & 276.6 & 0.80 & 0.61 \\
Faster-DiT      & 400  & 2.03 & 4.63   & 264.0 & 0.81 & 0.60 \\
DiT-XL/2        & 1400 & 2.27 & 4.60 & 278.2 & 0.83 & 0.57 \\
SiT-XL/2        & 1400 & 2.06 & 4.50 & 270.3 & 0.82 & 0.59 \\
\midrule
\multicolumn{7}{l}{\textit{Representation Alignment Methods (SiT-XL/2)}} \\
+ REPA        & 80   & 2.39 & 4.64 & 246.8 & 0.82 & 0.57 \\
+ REPA        & 200  & 1.96 & 4.49 & 264.0 & 0.82 & 0.60 \\
+ sREPA       & 80   & 2.25 & 4.65 & 257.5 & 0.83 & 0.58 \\
+ sREPA       & 200  & 1.91 & 4.50 & 271.8 & \textbf{0.83} & 0.60 \\
\rowcolor{repirow}
\textbf{+ REPI (ours)} & 80 &  2.22 & 4.58 & 236.4 & 0.81 & 0.59 \\ 
\rowcolor{repirow}
\textbf{+ REPI (ours)} & 200 & 1.90 & 4.49 & 255.1 & 0.81 & \textbf{0.61} \\ 
\rowcolor{combinedrow}
\textbf{+ REPA + REPI (ours)} & 80  & 2.08 & 4.55 & 245.7 & 0.81 & 0.59  \\
\rowcolor{combinedrow}
\textbf{+ REPA + REPI (ours)} & 200 &  \textbf{1.85} & \textbf{4.46} & \textbf{278.4} &  0.82 & 0.60  \\
\hdashline
\noalign{\vskip 2pt}
+ REPA*       & 80   & 1.98 & 4.60 & 263.0 & 0.80 & \textbf{0.61} \\
+ iREPA*      & 80   & 1.93 & 4.59 & 268.8 & 0.80 & 0.60 \\
+ Stable Velocity*  & 80   & 1.80 & \textbf{4.52} & 272.4 & \textbf{0.81} & 0.60 \\
\rowcolor{repirow}
\textbf{+ REPI (ours)}* & 80   & 1.79 & 4.61 &269.2 & 0.81 & 0.60 \\
\rowcolor{combinedrow}
\textbf{+ REPA + REPI (ours)}* & 80   & \textbf{1.73} & 4.62 & \textbf{278.4} & \textbf{0.81} & 0.60 \\
\bottomrule
\end{tabular}%
}

\endgroup
\end{minipage}

\vspace{-8pt}
\end{table*}

\section{Experiments}
\label{sec:experiments}

\subsection{Experimental Setup}
\label{sec:setup}

\textbf{Implementation details.}
We strictly follow the training protocols of REPA \citep{repa}. To ensure a fair comparison, we fix the training batch size to 256 and adopt the same learning rate and EMA configurations as REPA. For SiT \citep{sit} models, we employ the SDE Euler--Maruyama sampler with 250 function evaluations (NFE). When combined with REPA, we always use its default alignment depth and loss weight. Additional implementation details and hyperparameter settings are provided in Appendix~\ref{app:additional-details}.

\textbf{Models and datasets.}
We evaluate our method on DiT \citep{dit}, SiT \citep{sit}, and DiG \citep{DiG} across multiple model scales. Following REPA, our main experiments are conducted on class-conditional ImageNet \citep{imgnet} at $256\times256$ resolution. We further evaluate on ImageNet at $512\times512$ to assess higher-resolution generation, and on MS-COCO \citep{mscoco} with MM-DiT \citep{sd3} for text-to-image generation. Unless otherwise specified, we adopt DINOv2-B/14 \citep{dinov2} as the pretrained visual encoder, and all models operate in the latent space of a pretrained VAE \citep{vae}.

\textbf{Evaluation.}
We compare our method against four representation alignment methods: REPA~\citep{repa}, iREPA~\citep{irepa}, sREPA~\citep{xu2026beyond}, and Stable Velocity~\citep{stable_velocity}. We report FID~\citep{fid}, sFID~\citep{sfid}, IS~\citep{is}, and Precision/Recall~\citep{precrecall}, all computed on 50K generated samples following the ADM protocol~\citep{adm}. Unless otherwise noted, all results are obtained using the EMA model without classifier-free guidance (CFG)~\citep{cfg}. For our method, reported training steps account for the entire training process, from scaffold to internalization.

\subsection{Main Results}
\label{sec:main_results}

\textbf{Quantitative results.}
Table~\ref{tab:fid_comparison}, together with Figures~\ref{fig:teaser} and~\ref{fig:generality}, presents the quantitative comparison without classifier-free guidance. Across a diverse set of backbones, REPI consistently outperforms REPA, and combining the two methods yields further gains. Notably, on SiT-XL, REPA~+~REPI achieves an FID of 8.22 after only 160K training steps, matching the FID of 8.30 obtained by vanilla SiT after 7M steps, a training speedup of over $43.5\times$. Moreover, REPA~+~REPI at 200K steps already surpasses REPA at 400K steps. We further evaluate under classifier-free guidance, both with and without the guidance interval \citep{kynkaanniemi2024applying}. As shown in Table~\ref{tab:cfg_comparison}, REPI alone achieves comparable performance to the baselines, while REPA~+~REPI achieves the best FID and IS among all methods in both settings.

\begin{figure}[t]
    \centering
    \includegraphics[width=\linewidth]{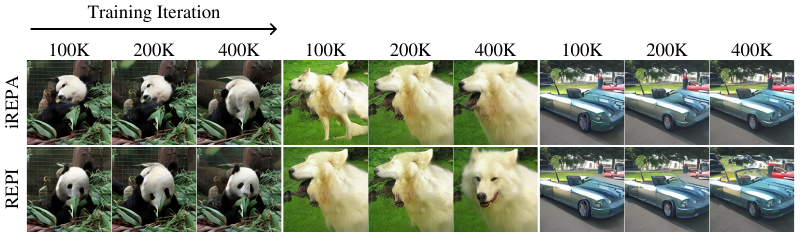}
    \caption{\textbf{Qualitative comparison across training iterations.} Generated samples from SiT-XL/2 trained with iREPA (top) and REPI (bottom) over the first 400K iterations. Both models use the same initial noise and sampler, without classifier-free guidance.}
    \label{fig:repa_repi_steps}
    \vspace{-10pt}
\end{figure}

\textbf{Qualitative results.} 
Figure~\ref{fig:repa_repi_steps} compares samples from our method and iREPA, generated from the same initial noise without classifier-free guidance. Our method produces images with more coherent semantic structure and finer details. Additional qualitative results are provided in Appendix~\ref{appendix:qualitative}.

\begin{wraptable}{r}{0.33\textwidth}
\vspace{-13.5pt}
    \centering
    \small
    \captionsetup{skip=3pt}
    \captionof{table}{FID comparison on text-to-image generation.}
    \label{tab:t2i}
    \setlength{\tabcolsep}{3.8pt}
    \renewcommand{\arraystretch}{1.08}
    \resizebox{1\linewidth}{!}{%
    \begin{tabular}{lcc}
        \toprule
        Method & w/o CFG & w/ CFG \\
        \midrule
        REPA & 10.40 & 4.73 \\
        \textbf{REPI} & 9.47 & 4.61 \\
        \textbf{REPA + REPI} & \textbf{9.08} & \textbf{4.56} \\
        \bottomrule
    \end{tabular}
}
\vspace{-8pt}
\end{wraptable}

\textbf{Text-to-image generation.}
Following REPA, we train MM-DiT on MS-COCO for 150K steps and use ODE sampling with NFE${}=50$, adopting the same evaluation protocol as REPA. As shown in Table~\ref{tab:t2i}, REPI consistently outperforms REPA both with and without CFG, reducing FID from 10.40 to 9.47 (w/o CFG) and 4.73 to 4.61 (w/ CFG). Combining REPI with REPA further reduces FID to 9.08 and 4.56, respectively.

\subsection{Generality of REPI}
\label{sec:generality}
In this section, we demonstrate that REPI improves diffusion transformer training across a broad range of settings, including visual encoders, diffusion backbones, and injection depths.

\begin{figure}[t]
    \centering
    \includegraphics[width=\linewidth]{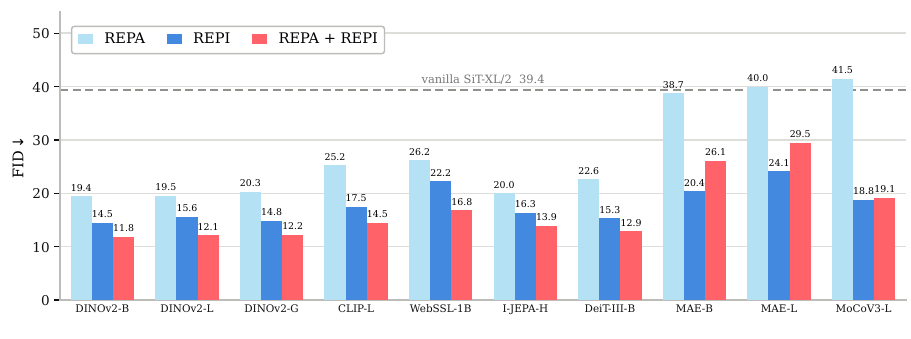}
    \caption{\textbf{Generality of REPI across visual encoders.} FID results of REPA, REPI, and REPA + REPI at 100K training steps on SiT-XL/2. REPI consistently outperforms REPA across all ten encoders. Notably, on encoders such as MAE-L and MoCoV3-L, where REPA underperforms vanilla SiT, REPI still achieves substantial gains.}
    \label{fig:encoder_fid_bar}
    \vspace{-6pt}
\end{figure}

\textbf{Visual encoders.}
We first study the effect of different visual encoder types and sizes. As shown in Figure~\ref{fig:encoder_fid_bar}, REPI consistently improves upon the vanilla model and outperforms REPA across all encoders. Notably, on several encoders (e.g., MAE and MoCoV3), REPA yields FID scores even worse than vanilla SiT, consistent with observations in iREPA \citep{irepa}. In contrast, REPI still achieves substantial gains on these encoders. For the remaining encoders, combining REPI with REPA further reduces FID substantially.

\begin{figure}[t]
    \centering
    \includegraphics[width=\linewidth]{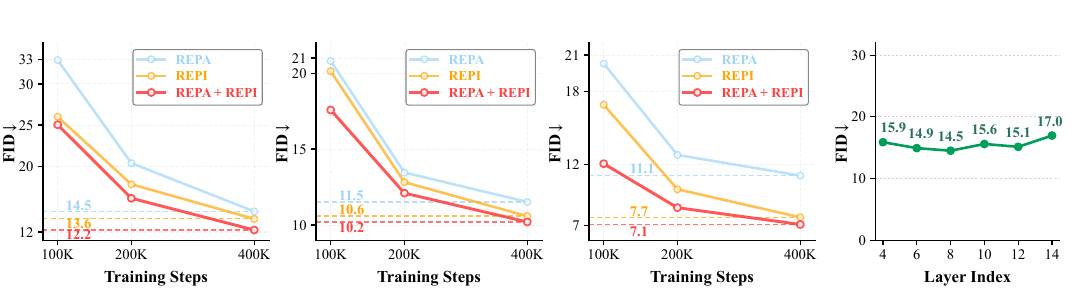}
    \par\vspace{-2pt}
    \noindent\makebox[\linewidth][l]{%
        \small
        \hspace*{6pt}%
        \hspace*{0.138343\linewidth}%
        \makebox[0pt][c]{(a) DiT-L/2}%
        \hspace*{0.251067\linewidth}%
        \makebox[0pt][c]{(b) DiG-L/2}%
        \hspace*{0.251067\linewidth}%
        \makebox[0pt][c]{(c) SiT-XL/2 (512)}%
        \hspace*{0.249289\linewidth}%
        \makebox[0pt][c]{(d) Injection depth}%
    }
    \caption{\textbf{Robustness of REPI across diffusion backbones and injection layers.} (a--c) FID comparison on DiT-L/2, DiG-L/2, and SiT-XL/2 ($512\times512$). REPI consistently outperforms REPA, and combining the two yields further gains. (d) REPI is robust to the choice of injection depth, with layer 8 achieving the best performance on SiT-XL/2 at 100K steps.}
    \label{fig:generality}
    \vspace{-7pt}
\end{figure}

\textbf{Diffusion backbones.}
We next evaluate REPI across a diverse set of diffusion transformer backbones. Figure~\ref{fig:teaser} reports results across model scales (SiT-B, SiT-L, and SiT-XL), while Figure~\ref{fig:generality} examines generality across architectures (DiT-L and DiG-L, spanning standard softmax attention and gated linear attention) and at a higher resolution (SiT-XL at $512\times512$). REPI generalizes consistently across model scales, architectures, and resolutions, outperforming REPA in every setting, and combining the two methods yields further substantial gains.

\textbf{Injection depth.}
Finally, we demonstrate the robustness of REPI to the choice of injection layer. As shown in Figure~\ref{fig:generality}(d), FID ranges only from 14.51 to 16.97 across injection layers 4 through 14, substantially below the 39.41 achieved by vanilla SiT-XL. Injecting at layer 8 achieves the best performance, consistent with the layer choice adopted in REPA.

\subsection{Ablation Studies}
\label{sec:ablations}
\begin{figure}[t]
    \centering
    \begin{minipage}[t]{0.526\linewidth}
        \centering
        \includegraphics[width=\linewidth]{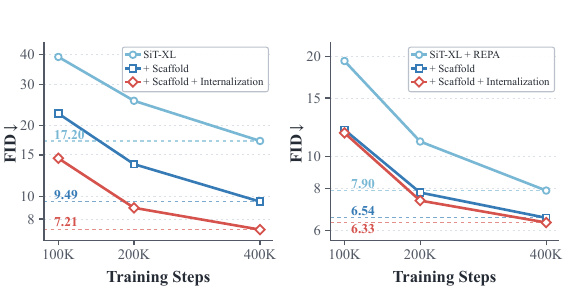}
        \par\vspace{-2pt}
        \footnotesize
        \makebox[\linewidth][l]{%
            \hspace*{0.245\linewidth}%
            \makebox[0pt][c]{(a) Without REPA}%
            \hspace*{0.525\linewidth}%
            \makebox[0pt][c]{(b) With REPA}%
        }
    \caption{\textbf{Ablation of the scaffold and internalization objective.} Starting from SiT-XL/2, (a) without REPA and (b) with REPA, we progressively add the scaffold and the internalization objective. Each component consistently improves FID.}
    
    \label{fig:component_ablations}
    \end{minipage}%
    \hspace{0.02\linewidth}%
    \begin{minipage}[t]{0.454\linewidth}
        \centering
        \includegraphics[width=\linewidth]{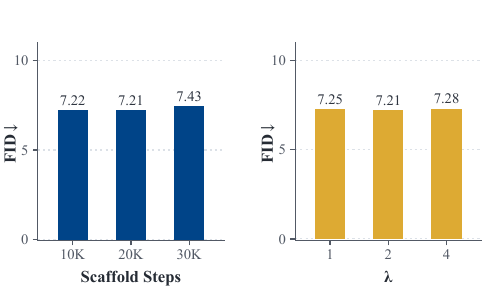}
        \par\vspace{-2pt}
        \footnotesize
        \makebox[\linewidth][l]{%
            \hspace*{0.247\linewidth}%
            \makebox[0pt][c]{(a) Scaffold duration}%
            \hspace*{0.538\linewidth}%
            \makebox[0pt][c]{(b) Loss weight}%
        }
        \captionof{figure}{\textbf{Robustness to scaffold duration and internalization loss weight.} REPI achieves stable FID across (a) scaffold durations and (b) internalization loss weights, evaluated on SiT-XL/2 at 400K training steps.}
        \label{fig:training_schedule_ablations}
    \end{minipage}
\end{figure}

\textbf{Scaffold and internalization.}
Figure~\ref{fig:component_ablations} ablates the contributions of the scaffold and internalization objective by progressively adding each component. The scaffold substantially improves performance over the base model, regardless of whether REPA is applied. Notably, using the scaffold alone, without REPA or the internalization objective (i.e., abrupt scaffold removal), reduces FID on SiT-XL from 17.2 to 9.49 at 400K steps. This suggests that the scaffold provides a favorable initialization for the diffusion transformer, with lasting benefits even after its removal. Applying the internalization objective further reduces FID, both with and without REPA.

\par\Needspace{9\baselineskip}
\begin{wraptable}{r}{0.31\textwidth}
\vspace{-12pt}
    \centering
    \small
    \captionof{table}{Injection-target ablation (SiT-XL/2, 100K steps).}
    \vspace{-8pt}
    \label{tab:ablation_injection_target}
    \setlength{\tabcolsep}{3.8pt}
    \renewcommand{\arraystretch}{1.08}
    \begin{tabular}{lrr}
        \toprule
        Target & FID$\downarrow$ & IS$\uparrow$ \\
        \midrule
        Attention output & 16.34 & 78.62 \\
        Q/K/V & 17.40 & 71.60 \\
        K only & 15.81 & 77.57 \\
        V only & 15.53 & 79.31 \\
        \rowcolor{gray!8}
        K/V & \textbf{14.51} & \textbf{82.71} \\
        \bottomrule
    \end{tabular}
\vspace{-8pt}
\end{wraptable}
\textbf{Injection target.} 
We next study where encoder representations should be injected within the attention module, comparing five variants summarized in Table~\ref{tab:ablation_injection_target}. All variants improve upon vanilla SiT, with K/V injection achieving the best FID and IS. Interestingly, this result contrasts with the oracle study in Section~\ref{sec:oracle}, where V-only injection performs best. We hypothesize that jointly injecting keys and values is better suited to scaffold removal and internalization: the projected keys determine how the native queries retrieve encoder-derived information, while the projected values supply the corresponding content. Notably, injecting into the attention output or into all of Q/K/V yields relatively worse results, suggesting that preserving native queries computed from the noisy input is important for internalization.

\textbf{Scaffold duration.}
We also investigate the effect of scaffold duration, varying it from 10K to 30K training steps. As shown in Figure~\ref{fig:training_schedule_ablations}(a), performance remains stable across durations, with FID varying by only 0.22, demonstrating that our method is robust to this choice. Notably, a short scaffold duration is sufficient to achieve strong performance in practice, as our oracle experiments show that the projection layers can be learned effectively within a small number of training steps.

\textbf{Internalization loss weight.}
Figure~\ref{fig:training_schedule_ablations}(b) shows that REPI is also robust to the internalization loss weight. Varying $\lambda$ from 1.0 to 4.0 yields similar FID scores, ranging between 7.21 and 7.28. We use $\lambda=2$ by default, which achieves the best result among the tested values.

We provide additional ablation studies for the combination of REPI and REPA in Appendix~\ref{sec:appendix_more_ablation}.

\section{Related Work}
\label{sec:related-work}

\textbf{Diffusion transformer.}
Diffusion Transformers (DiTs) \citep{dit} adopt Vision Transformers \citep{vit} as backbones for latent diffusion \citep{ldm}, while Scalable Interpolant Transformers (SiTs) \citep{sit} connect diffusion and flow matching \citep{fmgen} through stochastic interpolants. Subsequent works improve efficiency through linear attention \citep{SANA,DiG,wang2025litdelvingsimplifiedlinear,10.5555/3524938.3525416,wang2020linformer,choromanski2021rethinking} and masked image modeling \citep{zheng2024fast,Gao2023MaskedDT,MAE}, and training strategies that accelerate convergence without architectural modifications \citep{FasterDiT}.

\textbf{Representation alignment for generation.}
REPA \citep{repa} accelerates diffusion transformer training by aligning intermediate features with clean-image representations from pretrained visual encoders. Subsequent work refines the alignment target and training strategy by emphasizing spatial structure \citep{irepa}, aligning relational geometry \citep{xu2026beyond}, terminating alignment early in training \citep{wang2025repa}, or adopting more flexible and adaptive alignment schemes \citep{pang2026maskalign,wang2025learning,mo2026improving}. Other work derives alignment targets from the diffusion process itself \citep{sra,peng2026dual,chefer2026self} or from VAE features \citep{sra2,min2026ahpa}, while others optimize the generative latent space directly or jointly model image and semantic representations \citep{yao2025reconstruction,repae,kouzelis2025boosting,wu2025representation}. The paradigm has further been extended to U-Nets \citep{U-REPA}, pixel-space transformers \citep{shin2026representation}, and video generation \citep{zhang2025videorepa,wu2025geometry,lian2026sara}. Unlike these methods, which treat external representations as an alignment target, we assign them a complementary role: projected encoder representations are temporarily injected into the denoising computation as a scaffold, while the model's native representations gradually internalize this guidance.

\section{Conclusion}
\label{sec:conclusion}

In this paper, we introduced REPI, a training framework that leverages pretrained vision encoders in a direction complementary to REPA. Specifically, we investigated whether encoder representations can be injected into a diffusion transformer to benefit the denoising task. REPI follows a scaffold-to-internalization strategy: projected encoder representations initially serve as a temporary scaffold, while an internalization objective encourages the diffusion transformer to reproduce compatible representations on its own. Extensive experiments show that REPI improves both the generation quality and training efficiency of diffusion transformers. We hope this work will motivate further exploration of how external representations can be leveraged in generative training. Promising future directions include extending REPI beyond image generation to video generation and other generative domains.

\clearpage
\subsection*{AI Use Statement}
In this work, we used generative AI tools for language editing and polishing to improve the clarity and readability of the manuscript, as well as for code optimization and debugging. We did not use generative AI tools to generate research ideas, formulate hypotheses, design the methodology, conduct experiments, or interpret the results. All AI-assisted text was carefully reviewed and revised by the authors, and all AI-assisted code was verified and tested for correctness. We take full responsibility for the final content of this work, including any text, claims, or artifacts produced with the aid of generative AI.

\subsection*{Reproducibility Statement}
We provide detailed descriptions of the model architectures, training configurations, and hyperparameter settings in Section~\ref{sec:setup} and Appendix~\ref{app:additional-details}. Our source code will be made publicly available to facilitate reproducibility.
\par

\bibliography{conference}
\bibliographystyle{conference}
\clearpage
\appendix
\section{Additional Implementation Details and Hyperparameters}
\label{app:additional-details}

\textbf{More implementation details.}
We strictly follow the training protocol of REPA \citep{repa}. Latent representations are pre-computed using the \texttt{stabilityai/sd-vae-ft-ema} VAE encoder. The K/V projection heads $h_{\phi_K}$ and $h_{\phi_V}$ are each implemented as a single linear layer. All parameters are optimized with AdamW \citep{adamw}, using a constant learning rate of $10^{-4}$, momentum parameters $(\beta_1, \beta_2)\!=\!(0.9, 0.999)$, and no weight decay. Training is conducted in fp16 mixed precision with \texttt{torch.compile} for improved throughput, together with gradient clipping and an exponential moving average (EMA) of the generative model weights for stable optimization. For DiT-L/2 and DiG-L/2, we adopt the Improved DDPM objective of~\citet{iddpm}, predicting both noise and variance. Hyperparameters for different backbones when using REPI alone (i.e., without REPA) are summarized in Table~\ref{tab:hparams}.

\begin{table}[!htbp]
\centering
\small
\setlength{\tabcolsep}{4.9pt}
\renewcommand{\arraystretch}{1.12}
\captionsetup{skip=6pt}
\caption{\textbf{Hyperparameters and model configurations for standalone REPI}, shared across all ImageNet $256\times256$ and $512\times512$ experiments.}
\label{tab:hparams}
\begin{tabular}{p{0.25\linewidth}*{5}{>{\centering\arraybackslash}p{0.12\linewidth}}}
\toprule
 & SiT-B/2 & SiT-L/2 & SiT-XL/2 & DiT-L/2 & DiG-L/2 \\
\midrule
\textbf{Architecture} & & & & & \\
Number of layers & 12 & 24 & 28 & 24 & 24 \\
Hidden dimension & 768 & 1024 & 1152 & 1024 & 1024 \\
Number of heads & 12 & 16 & 16 & 16 & 16 \\
\midrule
\textbf{REPI} & & & & & \\
$\lambda$ & 2.0 & 2.0 & 2.0 & 2.0 & 4.0 \\
Injection layer & 6 & 8 & 8 & 8 & 8 \\
Scaffold duration & 20K & 20K & 20K & 20K & 20K \\
K/V projection layer & Linear & Linear & Linear & Linear & Linear \\
\midrule
\textbf{Optimization} & & & & & \\
Batch size & 256 & 256 & 256 & 256 & 256 \\
Optimizer & AdamW & AdamW & AdamW & AdamW & AdamW \\
Learning rate & $10^{-4}$ & $10^{-4}$ & $10^{-4}$ & $10^{-4}$ & $10^{-4}$ \\
$(\beta_1,\beta_2)$ & (0.9, 0.999) & (0.9, 0.999) & (0.9, 0.999) & (0.9, 0.999) & (0.9, 0.999) \\
Weight decay & 0 & 0 & 0 & 0 & 0 \\
\midrule
\textbf{Diffusion} & & & & & \\
Objective & \makecell{Linear\\interpolants} & \makecell{Linear\\interpolants} & \makecell{Linear\\interpolants} & \makecell{Improved\\DDPM} & \makecell{Improved\\DDPM} \\
Prediction & Velocity & Velocity & Velocity & \makecell{Noise and\\variance} & \makecell{Noise and\\variance} \\
Sampler & \makecell{Euler--\\Maruyama} & \makecell{Euler--\\Maruyama} & \makecell{Euler--\\Maruyama} & \makecell{Respaced\\DDPM} & \makecell{Respaced\\DDPM} \\
Sampling steps & 250 & 250 & 250 & 250 & 250 \\
\bottomrule
\end{tabular}
\vspace{-5pt}
\end{table}

\textbf{Combining REPI with REPA.}
When combining with REPA, all settings are kept consistent with those used for REPI alone (Table~\ref{tab:hparams}), except that the injection layer is changed to layer 4 and $\lambda$ is set to 0.25. For REPA, we adopt all of its default settings, including the alignment depth and alignment loss weight. Ablations for these combination-specific settings are reported in Appendix~\ref{sec:appendix_more_ablation}.

\textbf{Pretrained encoders.}
For our main results, we adopt DINOv2-B/14 as the pretrained visual encoder, following common practice in prior representation alignment work. In Section~\ref{sec:generality}, we further evaluate REPI's generality across a broad range of pretrained encoders, including MAE \citep{MAE}, MoCoV3 \citep{mocov3}, CLIP \citep{clip}, DINOv2 \citep{dinov2}, WebSSL \citep{fan2025scaling}, I-JEPA-H \citep{ijepa}, and DeiT-III \citep{touvron2022deit}.

\textbf{Training cost.}
All experiments are conducted on four NVIDIA H200 GPUs. Table~\ref{tab:training_cost} reports the total training time for 100K steps on SiT-XL/2. As shown, REPI introduces only marginal overhead over REPA (+3.2\% training time), while substantially improving generation quality, reducing FID by 25.2\% and improving IS by 22.7\%. Combining REPA and REPI further amplifies these gains, achieving a 39.3\% reduction in FID and a 43.6\% improvement in IS, at a modest additional cost of only 3.9\% training time over REPA.

\begin{table}[htbp]
    \centering
    \captionsetup{skip=6pt}
    \caption{\textbf{Total training time} for 100K steps on SiT-XL/2 using four NVIDIA H200 GPUs. Percentages in parentheses denote the relative change with respect to REPA.}    \label{tab:training_cost}
    \begin{tabular}{lccc}
        \toprule
        Method & Time (h) & FID$\downarrow$ & IS$\uparrow$ \\
\midrule
REPA & 4.63 & 19.40  & 67.4 \\
\midrule
REPI
& \makecell{4.78 \\ {\small(+3.2\%)}}
& \makecell{14.51 \\ {\small($-$25.2\%)}}
& \makecell{82.7 \\ {\small(+22.7\%)}} \\
\midrule
REPA + REPI
& \makecell{4.81 \\ {\small(+3.9\%)}}
& \makecell{11.78 \\ {\small($-$39.3\%)}}
& \makecell{96.8 \\ {\small(+43.6\%)}} \\
        \bottomrule
    \end{tabular}
\end{table}

\section{Details of Oracle Injection Variants}
\label{app:oracle}
\begin{figure}[bhtp]
    \centering
    \hspace*{-80pt}
    \includegraphics[width=0.5\linewidth]{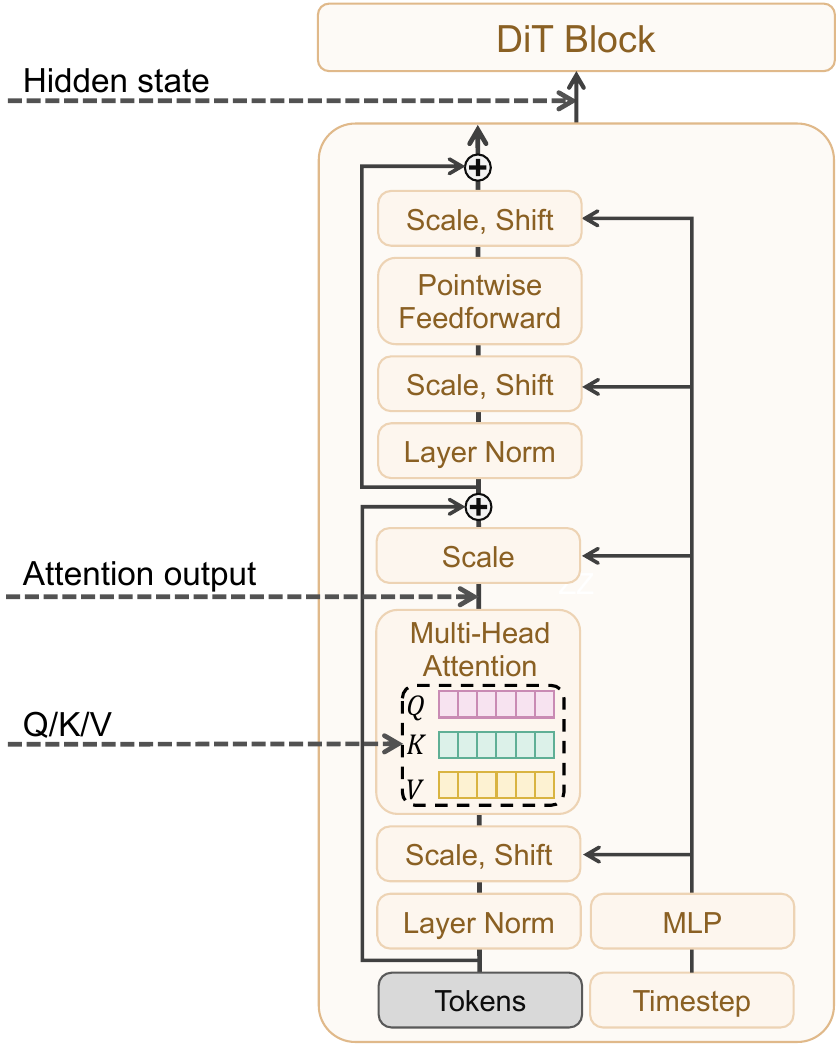}
    \caption{Illustration of injection targets within a DiT block.}
\label{fig:injection_interfaces}
\end{figure}

As described in Section~\ref{sec:oracle}, the oracle experiment is a diagnostic evaluation that assumes access to clean-image encoder representations at inference time. We investigate six injection targets: (1) hidden state, (2) attention output, (3) Q/K/V, (4) K/V, (5) K only, and (6) V only. The corresponding injection interfaces within a transformer block are illustrated in Figure~\ref{fig:injection_interfaces}. All six variants inject at block 8 of SiT-XL/2.

\textbf{Hidden state.}
The injection source is the encoder output, and the target is the hidden state of a transformer block (i.e., the block's output). As shown in Table~\ref{tab:oracle}, this setting fails catastrophically, yielding an FID of 212.5, because overwriting the block's output discards the noisy-input-dependent state accumulated by all preceding blocks.

\textbf{Attention output.}
The injection source is the encoder output, and the target is the attention output of a transformer block. As shown in Figure~\ref{fig:injection_interfaces}, this setting allows the noisy-input-dependent state to propagate to subsequent layers through the residual shortcut.

\textbf{Q/K/V.}
The injection source is the Q/K/V of the encoder's last attention layer, and the target is the corresponding Q/K/V of a transformer block. We evaluate four variants: Q/K/V, K only, V only, and K/V, each replacing the corresponding component(s). In all four variants, the noisy-input-dependent state can still propagate to subsequent layers through the residual shortcut, as well as through whichever Q/K/V components remain unreplaced.

\clearpage

\section{Additional Ablation Studies}
\label{sec:appendix_more_ablation}
We provide additional ablations of REPI when combined with REPA, examining the injection layer and internalization loss weight. Throughout these experiments, we retain REPA's default settings, including its alignment layer and alignment loss weight. We also report detailed metrics for the injection-target ablation, complementing Table~\ref{tab:ablation_injection_target} in the main paper.

\textbf{Injection layer.}
Table~\ref{tab:repi_repa_depth} reports results obtained by varying the REPI injection layer while fixing the REPA alignment layer at layer 8. Placing the injection layer before the alignment layer yields better performance than placing it after, with injection at layer 4 achieving the best result. As discussed in Section~\ref{sec:combining}, placing REPI first allows the subsequent REPA objective to supervise a representation that has already incorporated the injected encoder structure, encouraging this structure to propagate through subsequent layers. By contrast, injecting after alignment modifies the subsequent computation and may weaken the influence of the earlier alignment supervision.

\textbf{Internalization loss weight.}
Table~\ref{tab:repi_repa_lambda} reports ablations on the internalization loss weight $\lambda$ when combined with REPA. Our method remains robust to the choice of $\lambda$ in this setting: varying it from 0.125 to 1.0 yields similar FID scores, ranging between 11.78 and 12.68, substantially below vanilla SiT-XL's 39.4.

\textbf{Injection target.}
We further provide detailed results for the ablation on injection targets in Table~\ref{tab:ablation_injection}. All variants improve upon vanilla SiT, with K/V injection achieving the best performance.

\begin{table}[!htbp]
\centering
\setlength{\tabcolsep}{6pt}
\renewcommand{\arraystretch}{1.12}
\captionsetup{skip=6pt}
\caption{\textbf{Ablation results on the REPI injection layer when combined with REPA.} The REPA alignment layer is fixed at its default, layer 8. All results are reported on SiT-XL/2 at 100K training steps on ImageNet $256\times256$ without classifier-free guidance.}

\label{tab:repi_repa_depth}
  \begin{tabular}{cccccc}
  \toprule
   Layer & FID$\downarrow$ & sFID$\downarrow$ & IS$\uparrow$
        & Prec.$\uparrow$ & Rec.$\uparrow$ \\
  \midrule
  2  & 12.62 & 5.66 & 91.56 & 0.70 & 0.61 \\
  4  & \textbf{11.78} & \textbf{5.58} & \textbf{96.77}
     & \textbf{0.70} & \textbf{0.61} \\
  6  & 12.79 & 5.64 & 91.55 & 0.69 & 0.61 \\
  8  & 14.70 & 5.79 & 83.75 & 0.68 & 0.61 \\
  10 & 14.61 & 5.78 & 82.81 & 0.68 & 0.61 \\
  12 & 14.02 & 5.76 & 84.30 & 0.68 & 0.61 \\
  \bottomrule
  \end{tabular}
\end{table}

\begin{table}[!htbp]
\centering
\setlength{\tabcolsep}{6pt}
\renewcommand{\arraystretch}{1.12}
\captionsetup{skip=6pt}
\caption{\textbf{Ablation results on the internalization loss weight when combined with REPA.} All results are reported on SiT-XL/2 at 100K training steps on ImageNet $256\times256$ without classifier-free guidance.}
\label{tab:repi_repa_lambda}
\begin{tabular}{cccccc}
\toprule
$\lambda$ & FID$\downarrow$ & sFID$\downarrow$ & IS$\uparrow$ & Prec.$\uparrow$ & Rec.$\uparrow$ \\
\midrule
0.125 & 12.13 & 5.80 & 95.17 & 0.70 & 0.61 \\
0.25  & \textbf{11.78} & \textbf{5.58} & \textbf{96.77} & \textbf{0.70} & \textbf{0.61} \\
0.5   & 12.02 & 6.37 & 95.13 & 0.70 & 0.60 \\
1.0   & 12.68 & 7.05 & 93.18 & 0.69 & 0.60 \\
\bottomrule
\end{tabular}
\end{table}

\begin{table}[!htbp]
\centering
\setlength{\tabcolsep}{6pt}
\renewcommand{\arraystretch}{1.12}
\captionsetup{skip=6pt}
\caption{\textbf{Ablation results on the injection target.} All results are reported on SiT-XL/2 at 100K training steps on ImageNet $256\times256$ without classifier-free guidance.}
\label{tab:ablation_injection}
\begin{tabular}{lccccc}
\toprule
Injection target & FID$\downarrow$ & sFID$\downarrow$ & IS$\uparrow$
& Prec.$\uparrow$ & Rec.$\uparrow$ \\
\midrule
Attention output     & 16.34 & 5.89 & 78.62 & 0.67 & 0.60 \\
Q/K/V & 17.40 & 5.91 & 71.60 & 0.67 & 0.60 \\
K only            & 15.81 & 5.89 & 77.57 & 0.67 & 0.60 \\
V only          & 15.53 & \textbf{5.52} & 79.31 & 0.68 & 0.60 \\
K/V      & \textbf{14.51} & 5.57 & \textbf{82.71}
                     & \textbf{0.68} & \textbf{0.60} \\
\bottomrule
\end{tabular}
\end{table}

\clearpage
\section{Detailed Quantitative Results}
\label{appendix:quantitative}

Table~\ref{tab:detailed_results} provides more detailed quantitative results, complementing those in Table~\ref{tab:fid_comparison} of the main paper. All results are obtained using the same checkpoints and evaluation protocol, without classifier-free guidance. We compare REPA, REPI, and REPA~+~REPI at different training steps. REPI consistently outperforms REPA, and combining the two methods yields further improvements.

\begin{table}[h]
\centering
\captionsetup{skip=6pt}
\caption{\textbf{Detailed quantitative results across different SiT and DiT models.} All results are obtained on ImageNet $256\times256$ without classifier-free guidance.}
\label{tab:detailed_results}
\setlength{\tabcolsep}{5.8pt}
\begin{tabular}{lccccccc}
\toprule
Model & \#Params & Iter. & FID$\downarrow$ & sFID$\downarrow$ & IS$\uparrow$ & Prec.$\uparrow$ & Rec.$\uparrow$ \\
\midrule
\rowcolor{gray!15} SiT-B/2~\citep{sit} & 130M & 400K & 33.0 & 6.50 & 43.7 & 0.53 & 0.63 \\
\quad + REPA & 130M & 100K & 49.5 & 7.00 & 27.5 & 0.46 & 0.59 \\
\quad + REPA & 130M & 200K & 33.2 & 6.68 & 43.7 & 0.54 & 0.63 \\
\quad + REPA & 130M & 400K & 24.4 & 6.40 & 59.9 & 0.59 & 0.65 \\
\quad + REPI & 130M & 100K & 35.9 & 6.65 & 40.3 & 0.54 & 0.62 \\
\quad + REPI & 130M & 200K & 25.0 & 6.74 & 59.6 & 0.59 & 0.65 \\
\quad + REPI & 130M & 400K & 19.5 & 6.49 & 74.3 & 0.61 & 0.65 \\
\quad + REPA + REPI & 130M & 100K & 35.8 & 7.29 & 42.1 & 0.53 & 0.62 \\
\quad + REPA + REPI & 130M & 200K & 24.0 & 7.07 & 62.5 & 0.59 & 0.64 \\
\quad + REPA + REPI & 130M & 400K & 18.3 & 6.50 & 79.2 & 0.62 & 0.65 \\

\midrule
\rowcolor{gray!15} SiT-L/2~\citep{sit} & 458M & 400K & 18.8 & 5.30 & 72.0 & 0.64 & 0.64 \\
\quad + REPA & 458M & 100K & 24.1 & 6.25 & 55.7 & 0.62 & 0.60 \\
\quad + REPA & 458M & 200K & 14.0 & 5.18 & 86.5 & 0.67 & 0.64 \\
\quad + REPA & 458M & 400K & 9.7 & 5.20 & 109.2 & 0.69 & 0.65 \\
\quad + REPI & 458M & 100K & 19.0 & 5.48 & 67.8 & 0.65 & 0.62 \\
\quad + REPI & 458M & 200K & 12.1 & 5.16 & 96.2 & 0.68 & 0.64 \\
\quad + REPI & 458M & 400K & 9.0 & 5.10 & 115.5 & 0.69 & 0.65 \\
\quad + REPA + REPI & 458M & 100K & 14.7 & 5.63 & 84.9 & 0.67 & 0.62 \\
\quad + REPA + REPI & 458M & 200K & 10.1 & 5.21 & 109.2 & 0.69 & 0.64 \\
\quad + REPA + REPI & 458M & 400K & 8.2 & 5.26 & 124.4 & 0.69 & 0.65 \\

\midrule
\rowcolor{gray!15} SiT-XL/2~\citep{sit} & 675M & 7M & 8.3 & 6.30 & 131.7 & 0.68 & 0.67 \\
\quad + REPA & 675M & 100K & 19.4 & 6.06 & 67.4 & 0.64 & 0.61 \\
\quad + REPA & 675M & 200K & 11.1 & 5.05 & 100.4 & 0.69 & 0.64 \\
\quad + REPA & 675M & 400K & 7.9  & 5.06 & 122.6 & 0.70 & 0.65 \\
\quad + REPI & 675M & 100K & 14.5 & 5.57 & 82.7 & 0.68 & 0.60 \\
\quad + REPI & 675M & 200K & 8.9  & 4.96 & 113.3 & 0.70 & 0.63 \\
\quad + REPI & 675M & 400K & 7.2  & 5.03 & 131.4 & 0.70 & 0.66 \\
\quad + REPA + REPI & 675M & 100K & 11.8 & 5.58 & 96.8 & 0.70 & 0.61 \\
\quad + REPA + REPI & 675M & 160K & 8.2 & 4.90 & 118.9 & 0.71 & 0.63 \\
\quad + REPA + REPI & 675M & 200K & 7.4 & 4.86 & 127.3 & 0.71 & 0.64 \\
\quad + REPA + REPI & 675M & 400K & 6.3 & 5.02 & 140.7 & 0.71 & 0.66 \\

\midrule
\rowcolor{gray!15} DiT-L/2~\citep{dit} & 458M & 400K & 21.4 & 6.7 & 63.7 & 0.62 & 0.63 \\
\quad + REPA & 458M & 100K & 32.9 & 7.44 & 44.2 & 0.55 & 0.63 \\
\quad + REPA & 458M & 200K & 20.4 & 7.06 & 70.1 & 0.62 & 0.64 \\
\quad + REPA & 458M & 400K & 14.5 & 6.97 & 91.8 & 0.64 & 0.65 \\
\quad + REPI    & 458M & 100K & 26.0 & 6.44 & 53.3 & 0.60 & 0.62 \\
\quad + REPI    & 458M & 200K & 17.8 & 6.71 & 75.3 & 0.63 & 0.63 \\
\quad + REPI    & 458M & 400K & 13.6 & 6.52 & 93.3 & 0.66 & 0.65 \\
\quad + REPA + REPI & 458M & 100K & 25.0 & 7.16 & 57.9 & 0.60 & 0.63 \\
\quad + REPA + REPI & 458M & 200K & 16.1 & 6.81 & 84.3 & 0.64 & 0.64 \\
\quad + REPA + REPI & 458M & 400K & 12.2 & 6.70 & 101.7 & 0.66 & 0.66 \\

\bottomrule

\end{tabular}
\vspace{-10pt}
\end{table}

\clearpage
\section{Additional Qualitative Results}
\label{appendix:qualitative}
\begin{figure}[H]
  \centering
  \includegraphics[width=\linewidth]{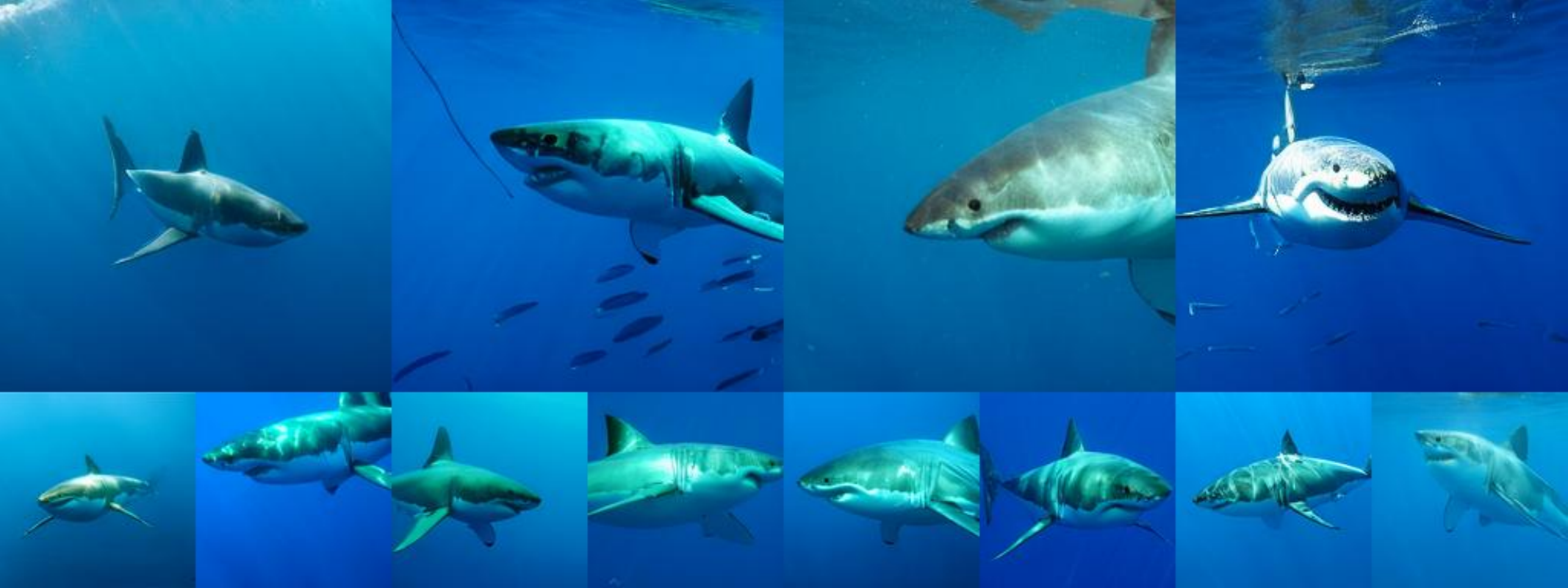}
  \caption{Generated samples from REPA~+~REPI on SiT-XL/2, trained for 400K steps, using classifier-free guidance ($w = 4.0$), conditioned on class ``great white shark'' (2).}
  \vspace{-10pt}
  \label{fig:qual_cls2}
\end{figure}

\begin{figure}[H]
  \centering
  \includegraphics[width=\linewidth]{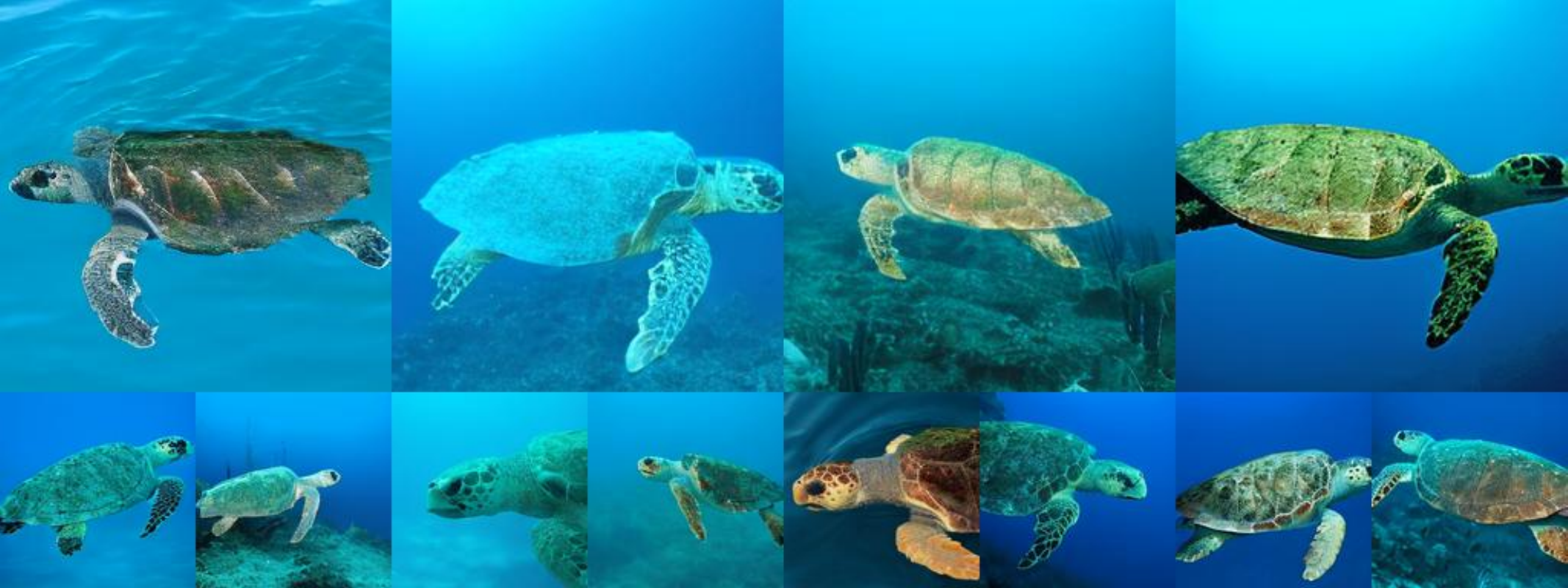}
  \caption{Generated samples from REPA~+~REPI on SiT-XL/2, trained for 400K steps, using classifier-free guidance ($w = 4.0$), conditioned on class ``loggerhead sea turtle'' (33).}
  \vspace{-10pt}
  \label{fig:qual_cls33}
\end{figure}

\begin{figure}[H]
  \centering
  \includegraphics[width=\linewidth]{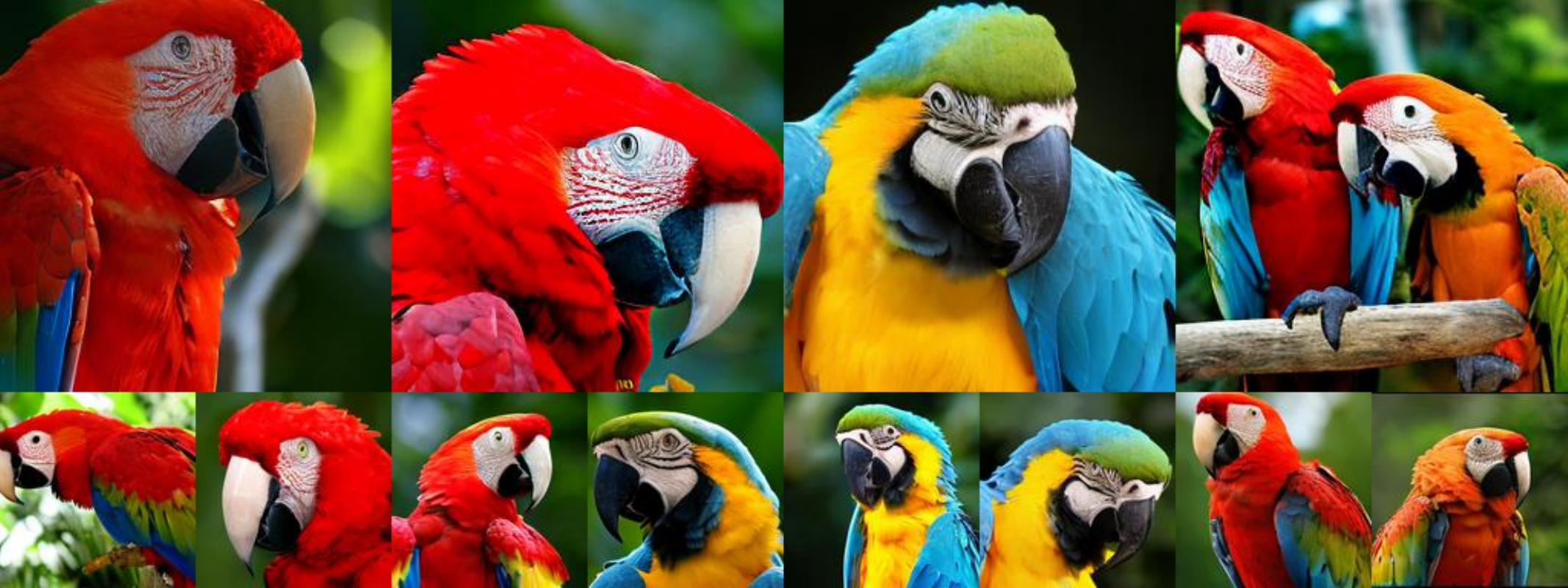}
  \caption{Generated samples from REPA~+~REPI on SiT-XL/2, trained for 400K steps, using classifier-free guidance ($w = 4.0$), conditioned on class ``macaw'' (88).}
  \label{fig:qual_cls88}
\end{figure}

\begin{figure}[H]
  \centering
  \includegraphics[width=\linewidth]{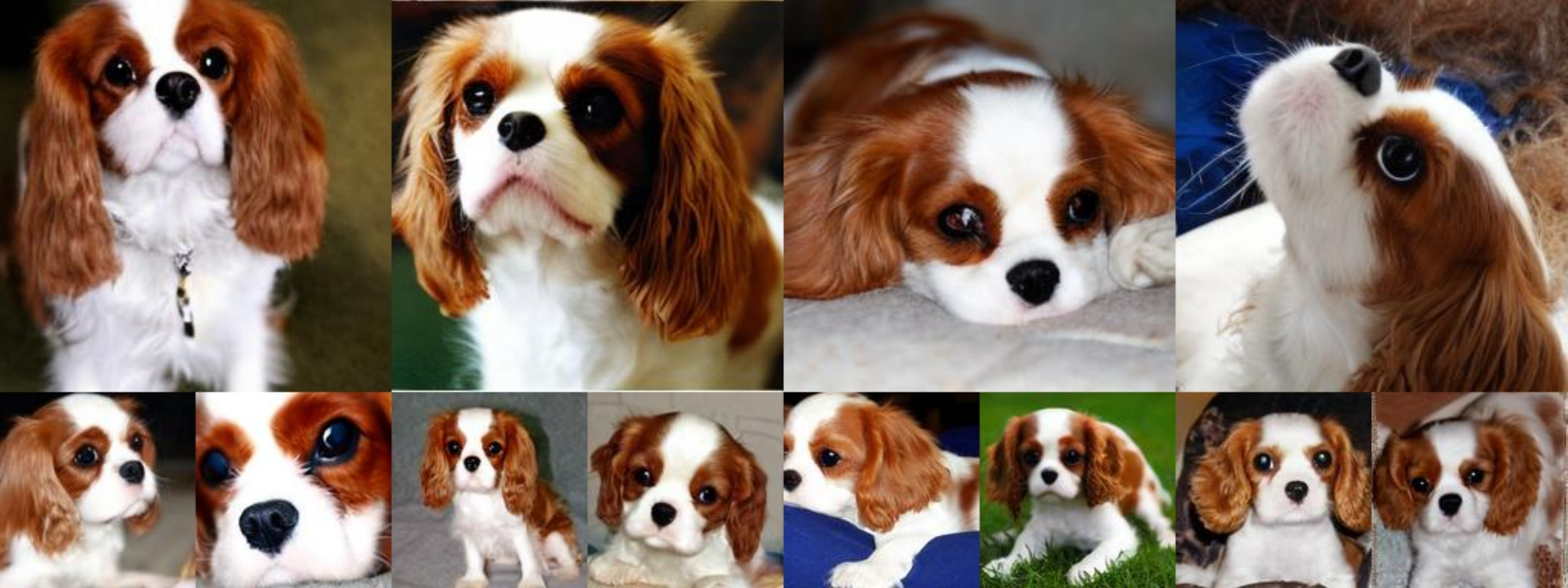}
  \caption{Generated samples from REPA~+~REPI on SiT-XL/2, trained for 400K steps, using classifier-free guidance ($w = 4.0$), conditioned on class ``Blenheim Spaniel'' (156).}
  \label{fig:qual_cls156}
\end{figure}

\begin{figure}[H]
  \centering
  \includegraphics[width=\linewidth]{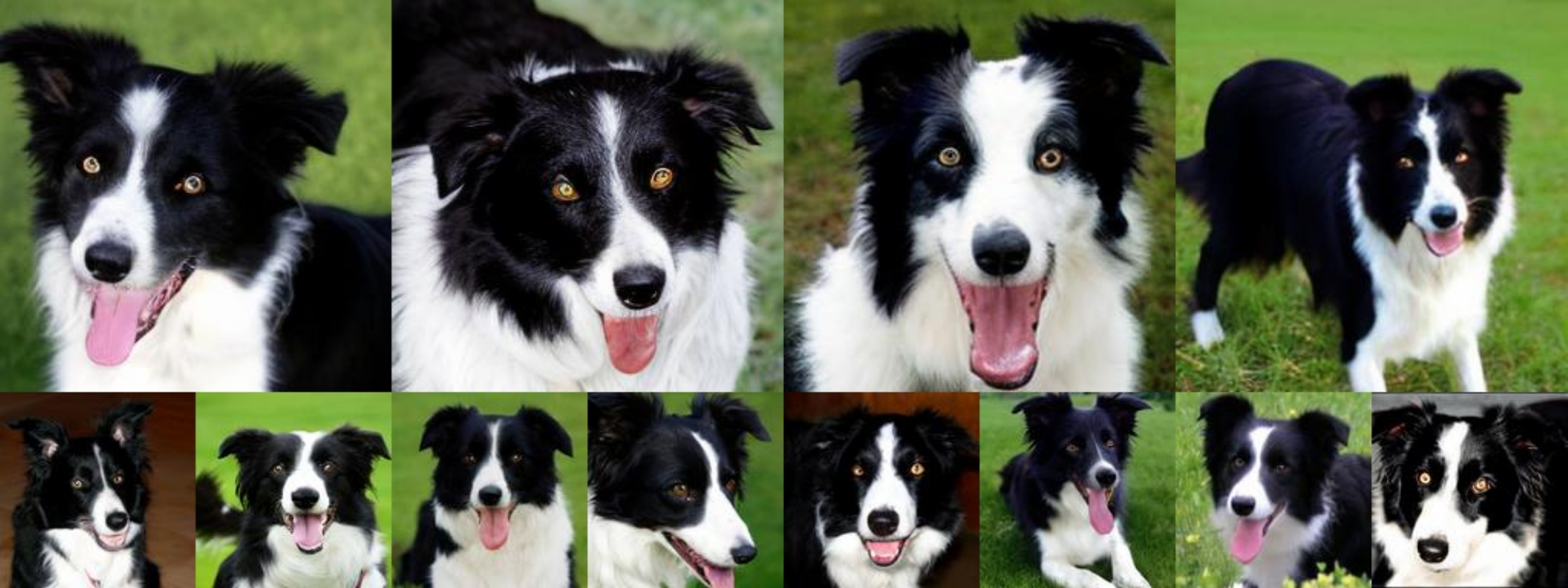}
  \caption{Generated samples from REPA~+~REPI on SiT-XL/2, trained for 400K steps, using classifier-free guidance ($w = 4.0$), conditioned on class ``Border collie'' (232).}
  \label{fig:qual_cls232}
\end{figure}

\begin{figure}[H]
  \centering
  \includegraphics[width=\linewidth]{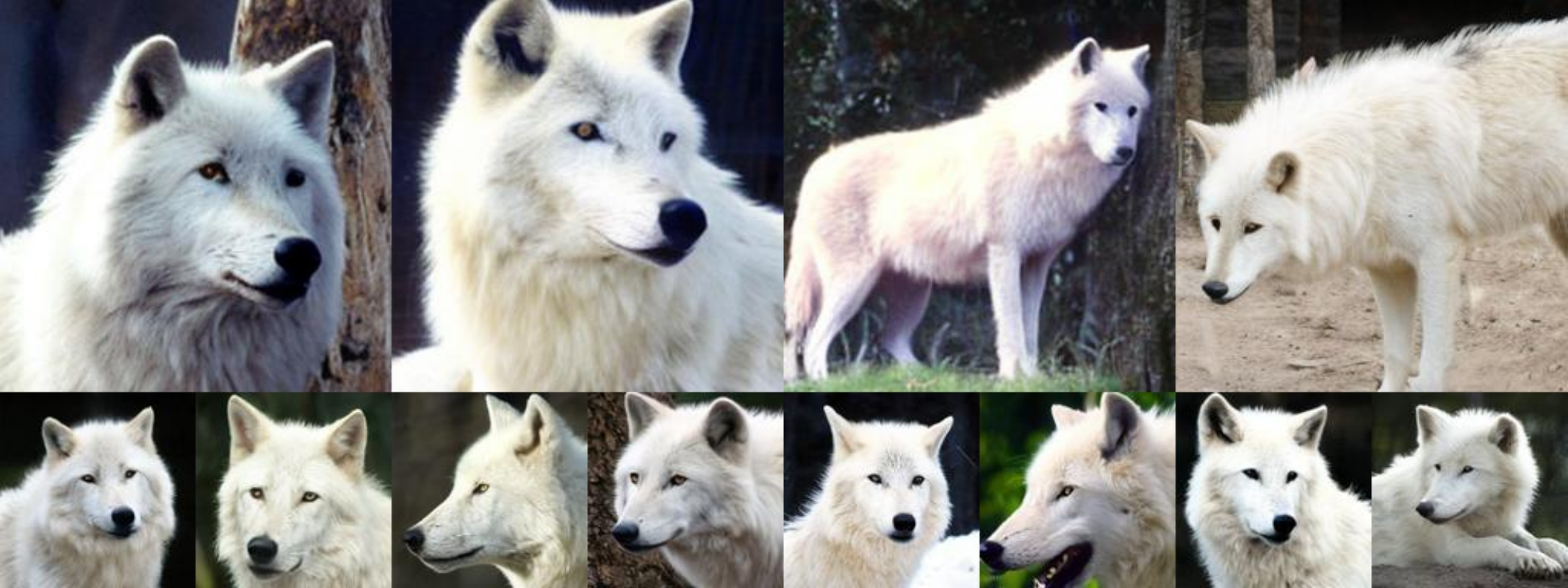}
  \caption{Generated samples from REPA~+~REPI on SiT-XL/2, trained for 400K steps, using classifier-free guidance ($w = 4.0$), conditioned on class ``Arctic wolf'' (270).}
  \label{fig:qual_cls270}
\end{figure}

\begin{figure}[H]
  \centering
  \includegraphics[width=\linewidth]{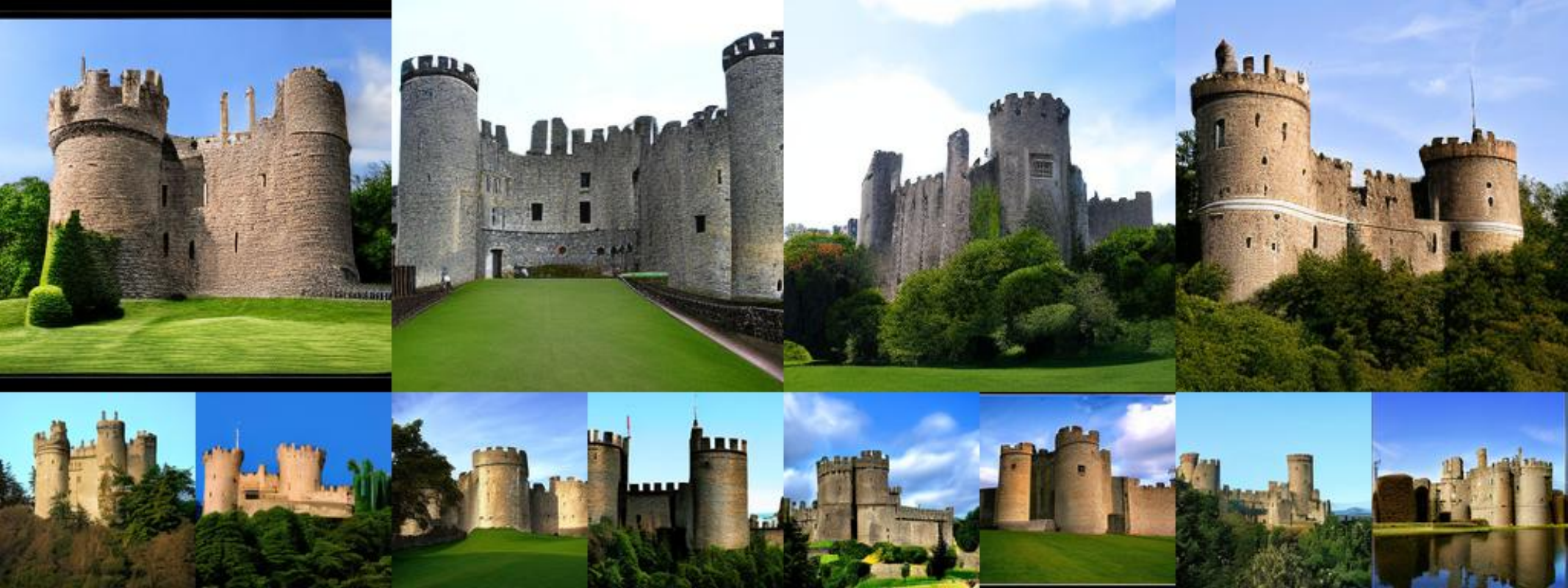}
  \caption{Generated samples from REPA~+~REPI on SiT-XL/2, trained for 400K steps, using classifier-free guidance ($w = 4.0$), conditioned on class ``castle'' (483).}
  \label{fig:qual_cls483}
\end{figure}

\begin{figure}[H]
  \centering
  \includegraphics[width=\linewidth]{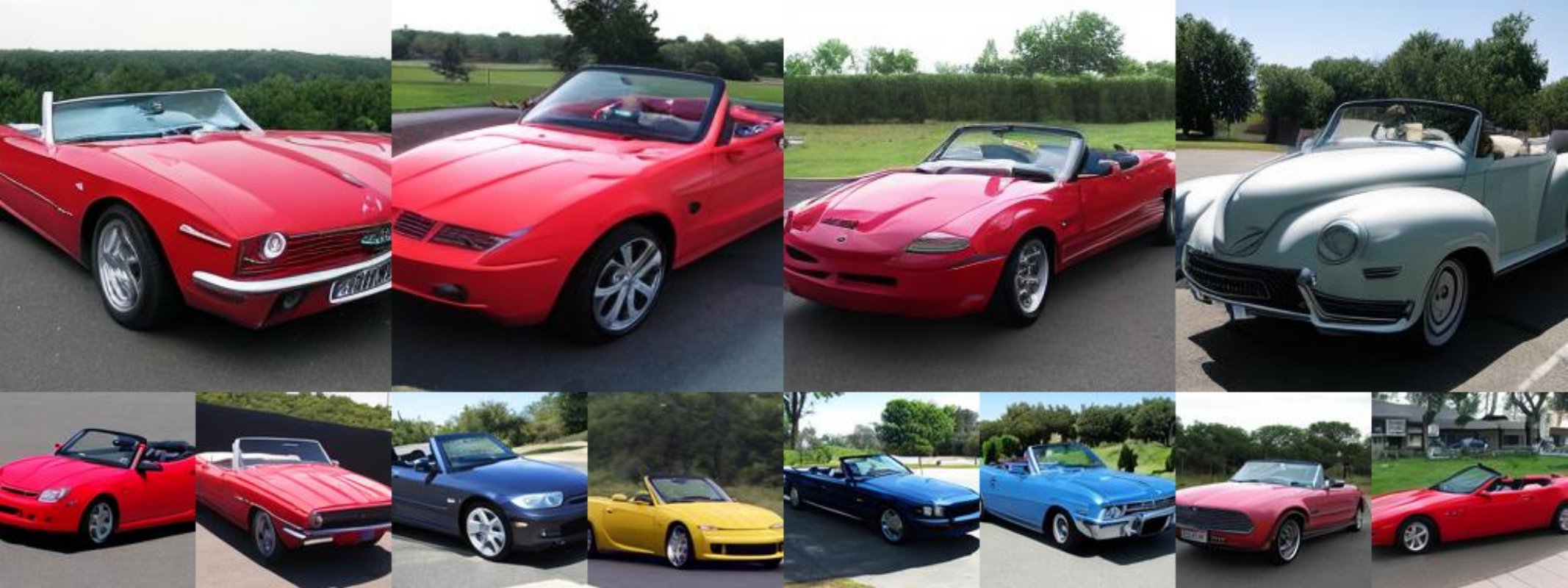}
  \caption{Generated samples from REPA~+~REPI on SiT-XL/2, trained for 400K steps, using classifier-free guidance ($w = 4.0$), conditioned on class ``convertible'' (511).}
  \label{fig:qual_cls511}
\end{figure}

\begin{figure}[H]
  \centering
  \includegraphics[width=\linewidth]{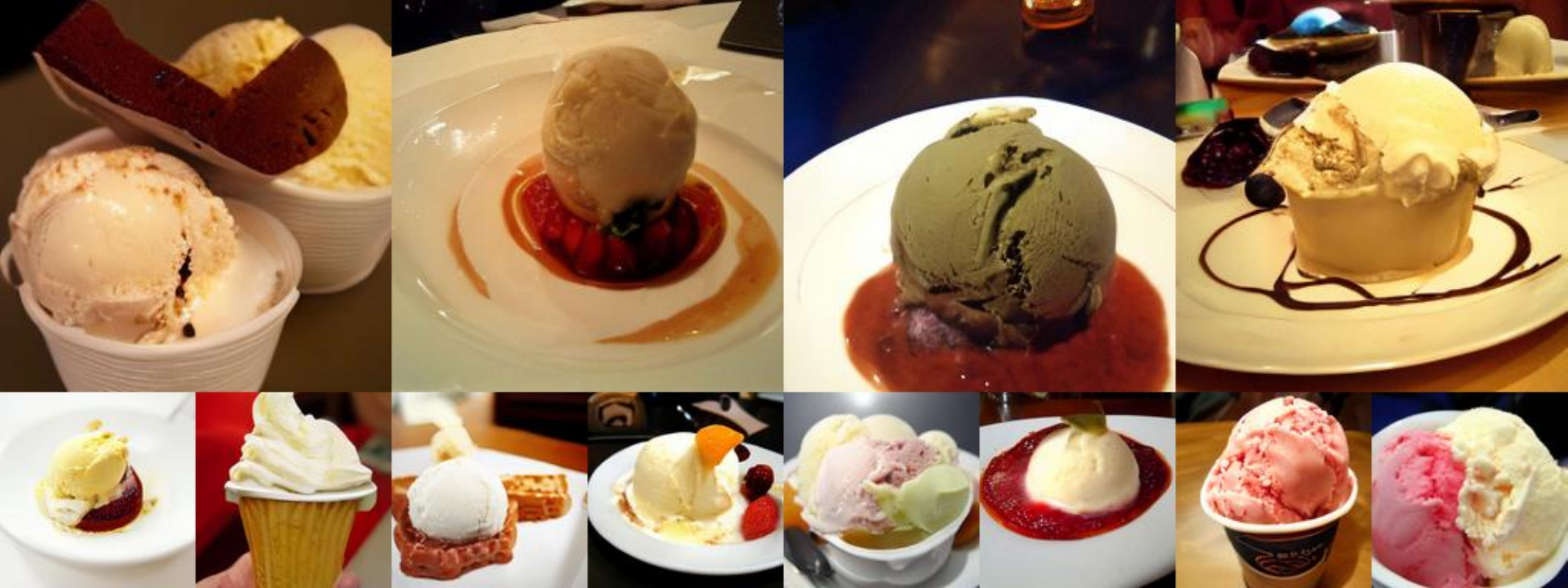}
  \caption{Generated samples from REPA~+~REPI on SiT-XL/2, trained for 400K steps, using classifier-free guidance ($w = 4.0$), conditioned on class ``ice cream'' (928).}
  \label{fig:qual_cls928}
\end{figure}

\begin{figure}[H]
  \centering
  \includegraphics[width=\linewidth]{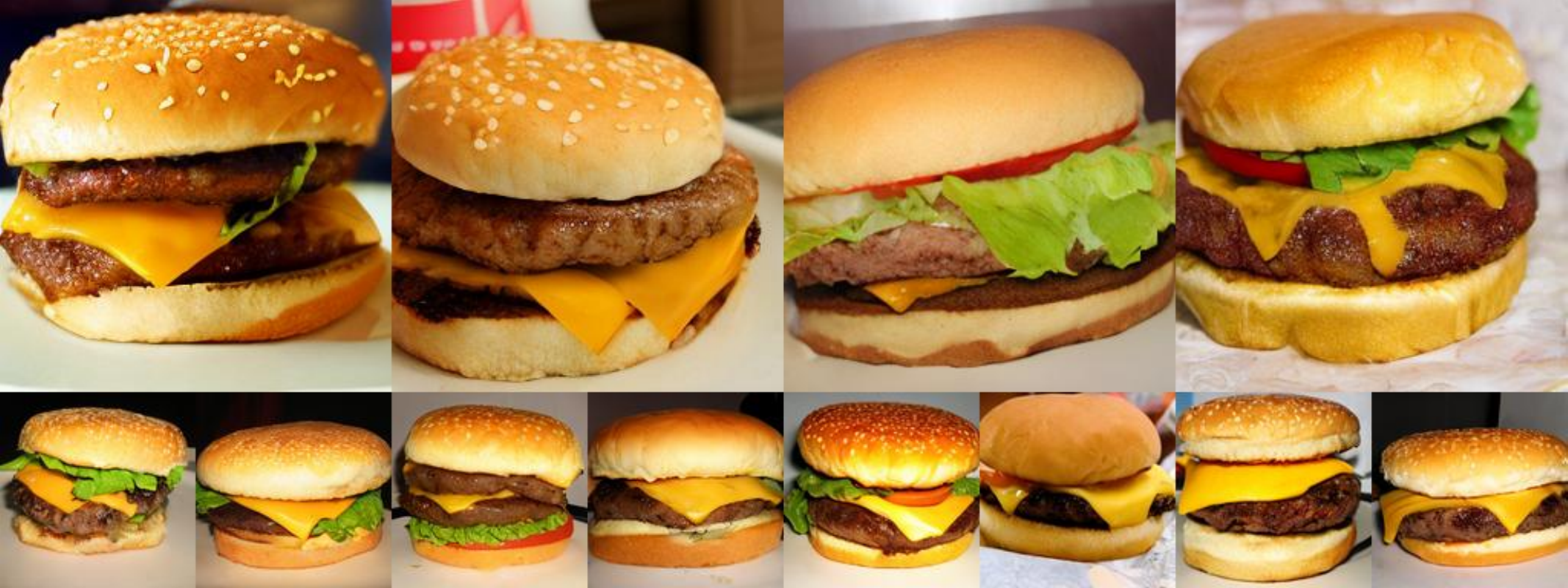}
  \caption{Generated samples from REPA~+~REPI on SiT-XL/2, trained for 400K steps, using classifier-free guidance ($w = 4.0$), conditioned on class ``cheeseburger'' (933).}
  \label{fig:qual_cls933}
\end{figure}

\begin{figure}[H]
  \centering
  \includegraphics[width=\linewidth]{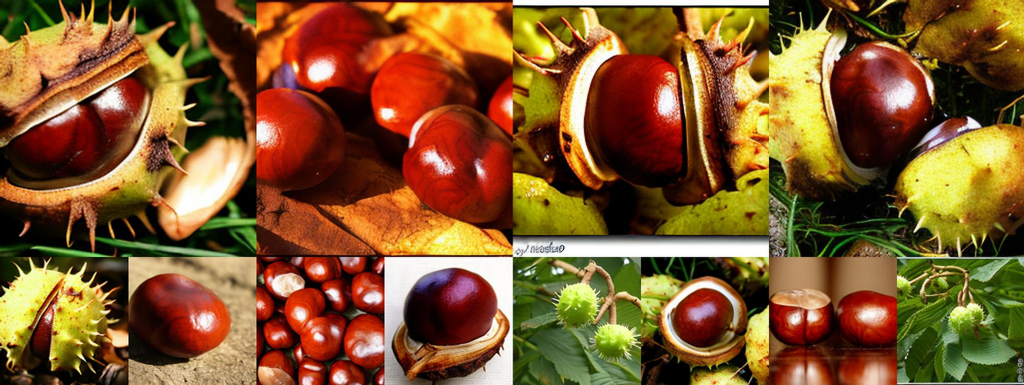}
  \caption{Generated samples from REPA~+~REPI on SiT-XL/2, trained for 400K steps, using classifier-free guidance ($w = 4.0$), conditioned on class ``buckeye'' (990).}
  \label{fig:qual_cls990}
\end{figure}

\begin{figure}[H]
  \centering
  \includegraphics[width=\linewidth]{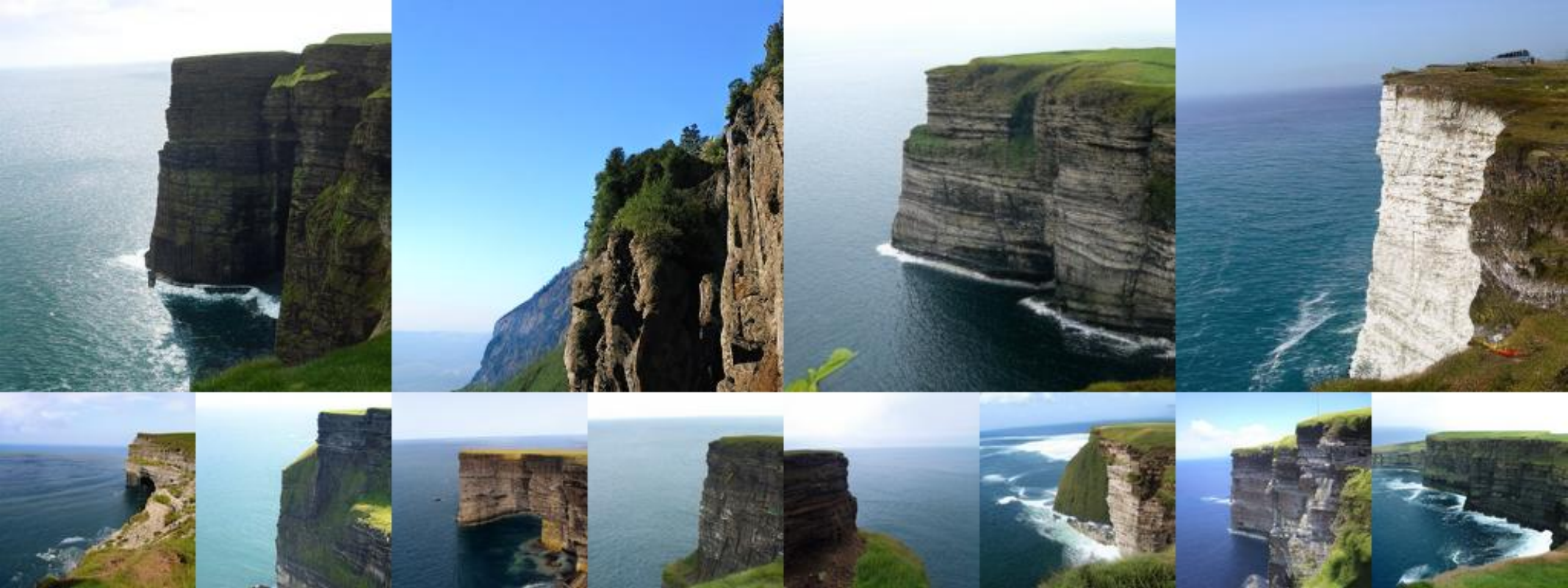}
  \caption{Generated samples from REPA~+~REPI on SiT-XL/2, trained for 400K steps, using classifier-free guidance ($w = 4.0$), conditioned on class ``cliff'' (972).}
  \label{fig:qual_cls972}
\end{figure}

\begin{figure}[H]
  \centering
  \includegraphics[width=\linewidth]{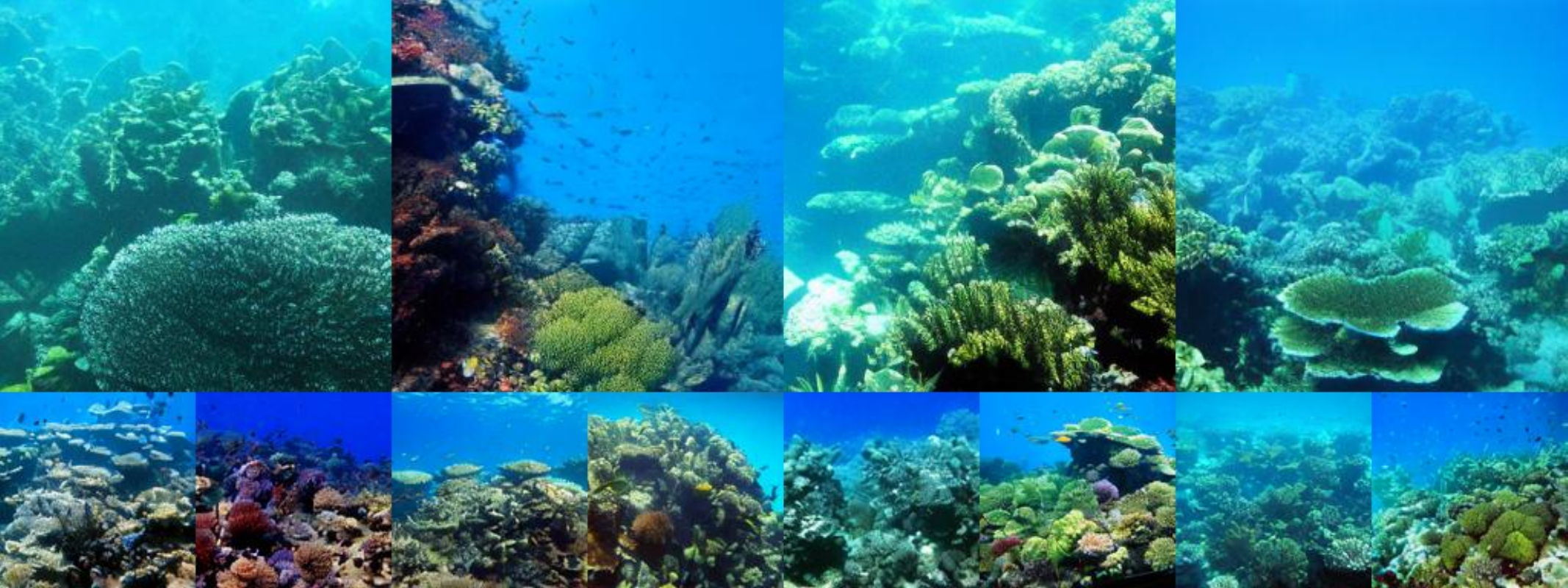}
  \caption{Generated samples from REPA~+~REPI on SiT-XL/2, trained for 400K steps, using classifier-free guidance ($w = 4.0$), conditioned on class ``coral reef'' (973).}
  \label{fig:qual_cls973}
\end{figure}

\begin{figure}[H]
  \centering
  \includegraphics[width=\linewidth]{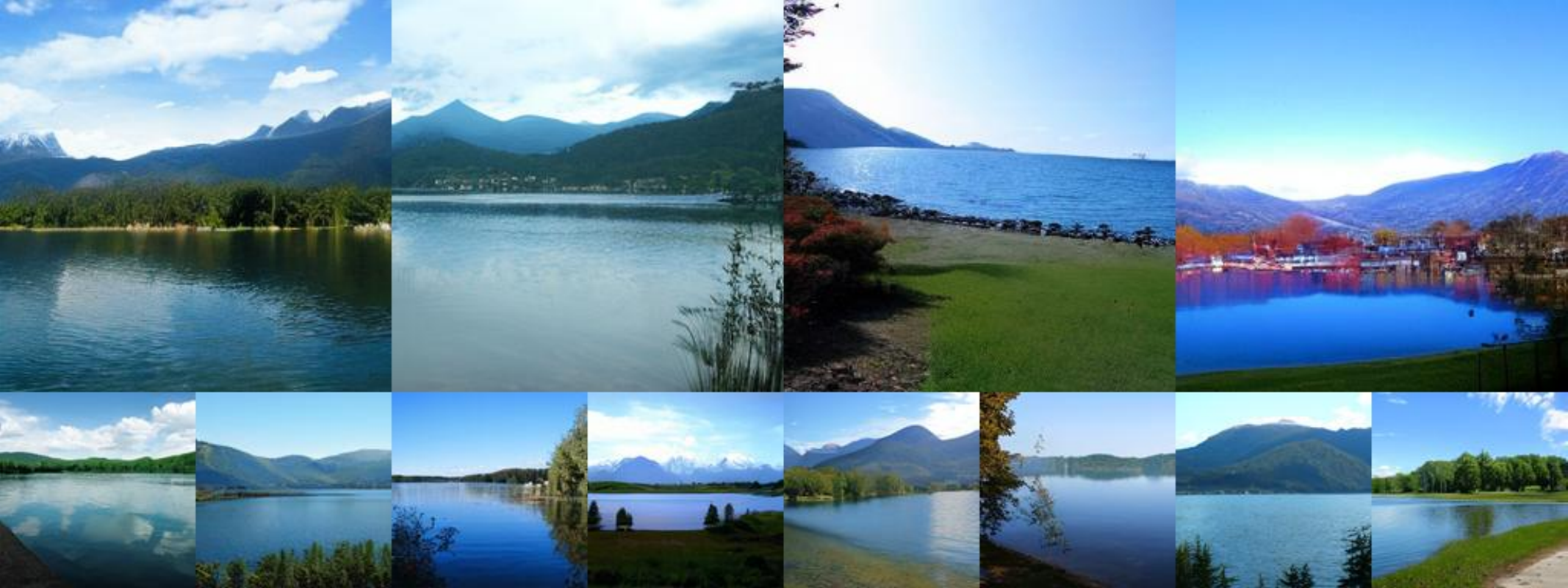}
  \caption{Generated samples from REPA~+~REPI on SiT-XL/2, trained for 400K steps, using classifier-free guidance ($w = 4.0$), conditioned on class ``lakeshore'' (975).}
  \label{fig:qual_cls975}
\end{figure}

\begin{figure}[H]
  \centering
  \includegraphics[width=\linewidth]{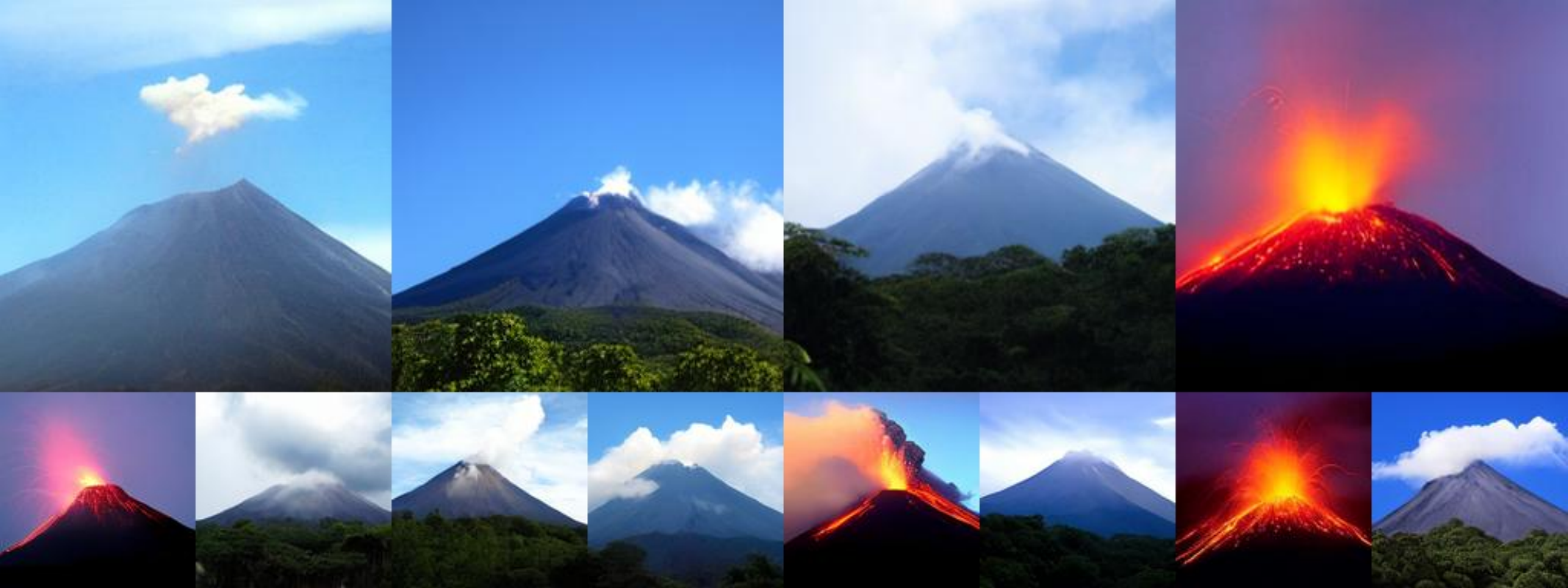}
  \caption{Generated samples from REPA~+~REPI on SiT-XL/2, trained for 400K steps, using classifier-free guidance ($w = 4.0$), conditioned on class ``volcano'' (980).}
  \label{fig:qual_cls980}
\end{figure}

\begin{figure}[p]
  \centering
  \includegraphics[width=\linewidth]{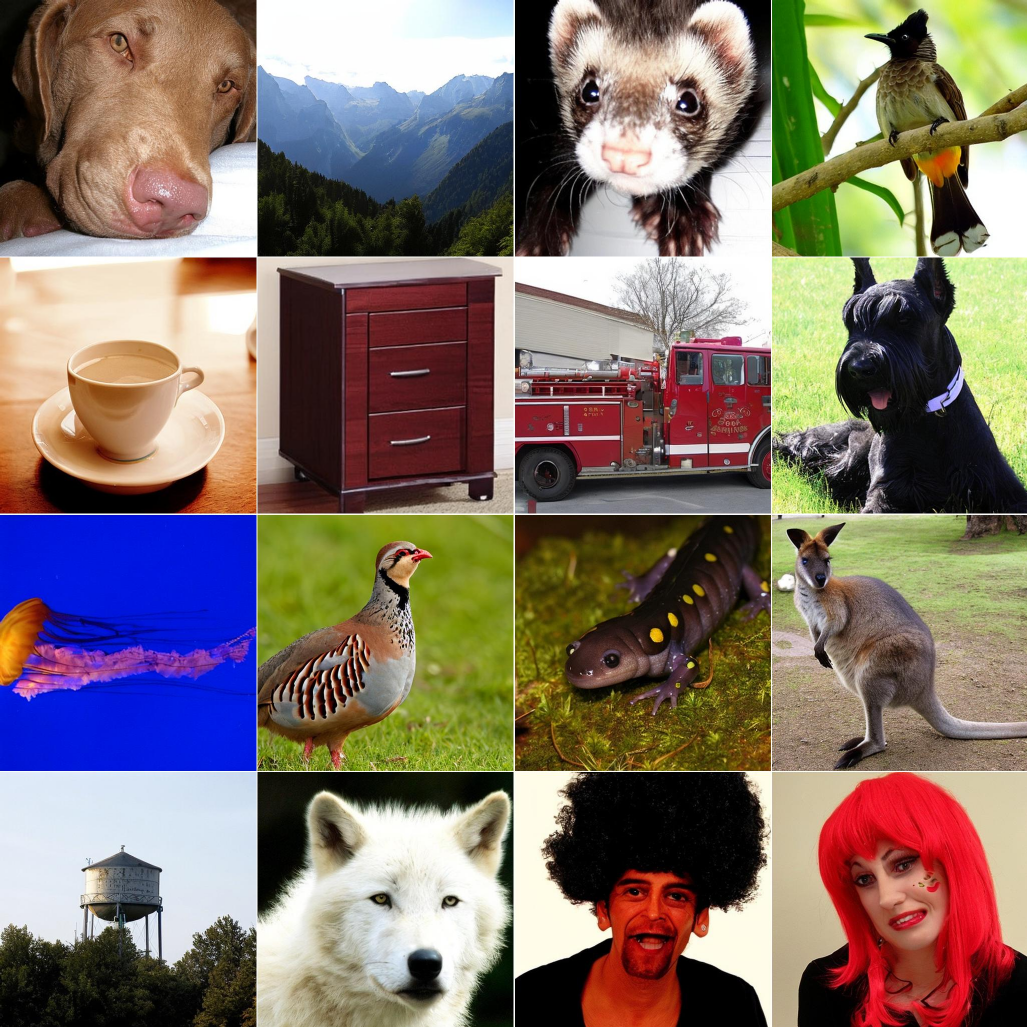}
\caption{Generated samples ($512\!\times\!512$) from REPA~+~REPI on SiT-XL/2, trained for 400K steps, using classifier-free guidance ($w = 4.0$).}
  \label{fig:512_qualitative}
\end{figure}

Figures~\ref{fig:qual_cls2}--\ref{fig:qual_cls980} present additional qualitative samples generated by REPA~+~REPI on SiT-XL/2, trained for 400K steps. Figure~\ref{fig:512_qualitative} presents additional samples at $512\times512$ resolution. All samples are generated with classifier-free guidance at scale $w = 4.0$.

\end{document}